\documentclass[11pt, letterpaper]{template}
\usepackage{xcolor}         % colors
\usepackage{lmodern} % for scalable font support
\usepackage[authoryear, round]{natbib}
\usepackage{hyperref}[citecolor=magenta]
\usepackage{url} 
\hypersetup{
    colorlinks=true,
    citecolor=blue,
    linkcolor=magenta,
    urlcolor=magenta
}

\definecolor{darkblue}{rgb}{0, 0, 0.5}
\usepackage{url}            % simple URL typesetting
\usepackage{booktabs}       % professional-quality tables
\usepackage{amsfonts}       % blackboard math symbols
\usepackage{nicefrac}       % compact symbols for 1/2, etc.
\usepackage{microtype}      % microtypography

\usepackage{lineno}

\usepackage{enumitem}       % [leftmargin=20pt]
\usepackage{subfig}         % subfloat
\usepackage{multirow}       % 
\usepackage{xspace}         % xspace
\usepackage{wrapfig}        % wrapfigure
\usepackage{makecell}       % divide row in table
\usepackage{colortbl}
\usepackage{bbm}            % indicator function (\mathbbm)
\usepackage{adjustbox}

\usepackage[most,skins,theorems]{tcolorbox}
\tcbset{
  aibox/.style={
    width=\linewidth,
    top=8pt,
    bottom=4pt,
    colback=blue!6!white,
    colframe=black,
    colbacktitle=black,
    enhanced,
    center,
    attach boxed title to top left={yshift=-0.1in,xshift=0.15in},
    boxed title style={boxrule=0pt,colframe=white,},
  }
}
\newtcolorbox{AIbox}[2][]{aibox,title=#2,#1}

\newcommand\blfootnote[1]{%
  \begingroup
  \renewcommand\thefootnote{}\footnote{#1}%
  \addtocounter{footnote}{-1}%
  \endgroup
}

\title{Resurrect Mask AutoRegressive Modeling for Efficient and Scalable Image Generation}

\author[1,2,5]{Yi Xin}
\author[2,3]{Le Zhuo}
\author[2]{Qi Qin}
\author[2,4]{Siqi Luo}
\author[2]{Yuewen Cao}
\author[2]{Bin Fu}
\author[6]{Yangfan He}
\author[3]{Hongsheng Li}
\author[4]{Guangtao Zhai}
\author[4,$\dag$]{Xiaohong Liu}
\author[2,$\dag$]{Peng Gao}

\affil[1]{Nanjing University}
\affil[2]{Shanghai AI Laboratory}
\affil[3]{The Chinese University of Hong Kong}
\affil[4]{Shanghai Jiao Tong University}
\affil[5]{Shanghai Innovation Institute}
\affil[6]{University of Minnesota Twin Cities}

\begin{abstract}
AutoRegressive (AR) models have made notable progress in image generation, with Masked AutoRegressive (MAR) models gaining attention for their efficient parallel decoding. However, MAR models have traditionally underperformed when compared to standard AR models. This study refines the MAR architecture to improve image generation quality. We begin by evaluating various image tokenizers to identify the most effective one. Subsequently, we introduce an improved Bidirectional LLaMA architecture by replacing causal attention with bidirectional attention and incorporating 2D RoPE, which together form our advanced model, MaskGIL. Scaled from 111M to 1.4B parameters, MaskGIL achieves a FID score of 3.71, matching state-of-the-art AR models in the ImageNet 256x256 benchmark, while requiring only 8 inference steps compared to the 256 steps of AR models. Furthermore, we develop a text-driven MaskGIL model with 775M parameters for generating images from text at various resolutions. Beyond image generation, MaskGIL extends to accelerate AR-based generation and enable real-time speech-to-image conversion. Our codes and models are available at \url{https://github.com/synbol/MaskGIL}.
\end{abstract}
\begin{document}

\blfootnote{$^\dag$ Corresponding author.}

\maketitle

\begin{figure*}[!t]
    \centering
    \includegraphics[width=1.0\linewidth]{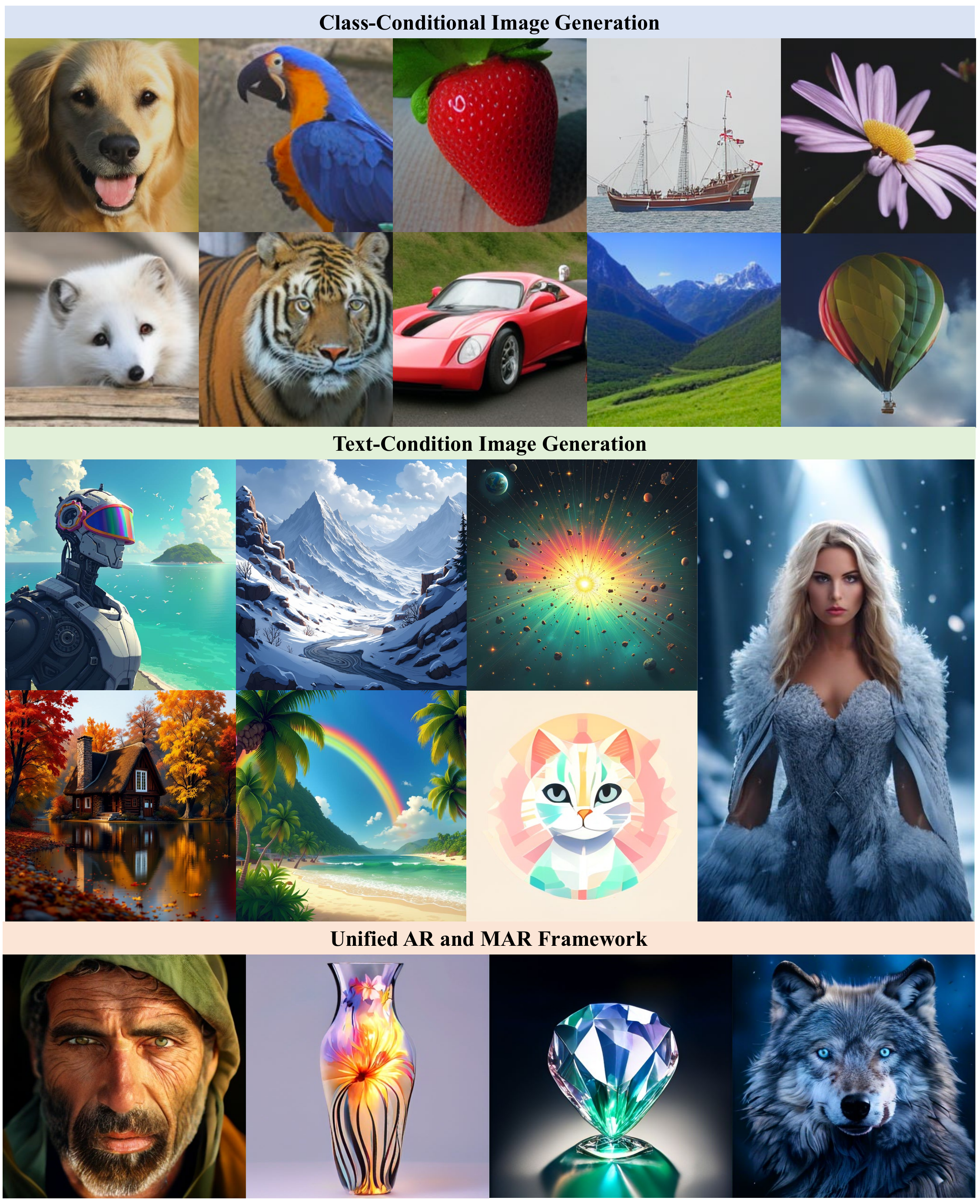}
    \caption{Image Generation with our MaskGIL Model and Unified Inference Framework. We show samples from our class-driven generation (top) and text-driven generation (middle) in various resolutions. At the bottom, we show the generation results of MaskGIL fused with Lumina-mGPT.}
    \label{fig:abs}
\end{figure*}

\section{Introduction}
AutoRegressive (AR) generative models have garnered increasing attention in image generation~\citep{vavqe2,vqgan,ramesh2021zero,lee2022autoregressive,parti,llamagen,liu2024lumina,liu2024customize,bao2025text,lumina-mgpt-2.0}, inspired by the success of Transformer~\citep{vaswani2017attention} and GPT~\citep{brown2020language,ChatGPT,gpt4} in NLP. This paradigm typically unfolds in two stages: the first stage is to quantize an image to a sequence of discrete tokens. In the second stage, an AR model is trained to predict the next token sequentially, based on the previously generated tokens. While AR models exhibit strong generative capabilities, their efficiency is hindered by the extensive number of inference steps, as the AR model \textit{\textbf{generates one token at a time}}. 

To overcome this limitation, Mask AutoRegressive (MAR) generative models~\citep{chang2022maskgit,lezama2022improved,li2023mage,qian2023strait,muse} have been developed, aiming to deliver high-quality image generation with fewer inference steps. Unlike traditional AR models that predict the next token, MAR models \textit{\textbf{predict a subset of tokens}}, offering a potential speed advantage. However, due to the inherent difficulties in training and prediction, the generative capabilities of existing MAR models remain less robust compared to AR models.

In this work, we aim to resurrect Mask Autoregressive (MAR) models for efficient image generation by systematically exploring their capability limits. We begin with an in-depth study to determine the optimal MAR architecture, considering both the choice of image tokenizer and the bidirectional model architecture. Among four prevalent tokenizers, including MaskGIT-VQ~\citep{chang2022maskgit}, Chameleon-VQ~\citep{chameleon}, LlamaGen-VQ~\citep{llamagen}, and Open-MAGVIT2-VQ~\citep{luo2024open-magvit2}, we find that LlamaGen-VQ with a codebook size of 16,384 achieves the best performance in MAR-based image generation. Regarding model architecture, inspired by the success of LLaMA~\citep{llama} in AR generation, we introduce an enhanced Bidirectional LLaMA architecture for MAR, named \textbf{Mask}ed \textbf{G}enerative \textbf{I}mage \textbf{L}LaMA (\textbf{MaskGIL}). This enhancement involves modifying causal attention with bidirectional attention and integrating 2D RoPE into every layer of the LLaMA architecture. Through comparative analysis, we find that MaskGIL demonstrates superior image generation capabilities.  

Building upon this architecture, we present a series of class-driven image generation models to explore the upper limits of model scaling, ranging from 111M to 1.4B parameters. However, we find that scaling models beyond 1.4B parameters presents significant challenges in maintaining stable training. To address this issue, we incorporate query-key normalization (QK-Norm) and Post-Norm, which effectively stabilize training at larger scales. Furthermore, we introduce a text-driven image generation model with 775M parameters, capable of flexibly generating images at various resolutions while maintaining high quality.

In addition to focusing on the fundamental image generation capabilities of MaskGIL, we have also expanded MaskGIL to broader applications, including: (1) Accelerating AR Generation: We propose a hybrid framework that leverages an AR model to generate a portion of the tokens, which then serve as prompts for MaskGIL to complete the remaining tokens. This hybrid approach significantly reduces the number of AR sampling steps while maintaining high visual fidelity. (2) Real-Time Speech-To-Image Generation System: The fast sampling strategy of MaskGIL makes it more suitable for interactive generation with users compared to pure AR generation. Therefore, we develop a real-time speech-to-image generation system based on our MaskGIL model. In summary, our contributions include:
\begin{figure}[!t]
    \centering
    \includegraphics[width=1.0\linewidth]{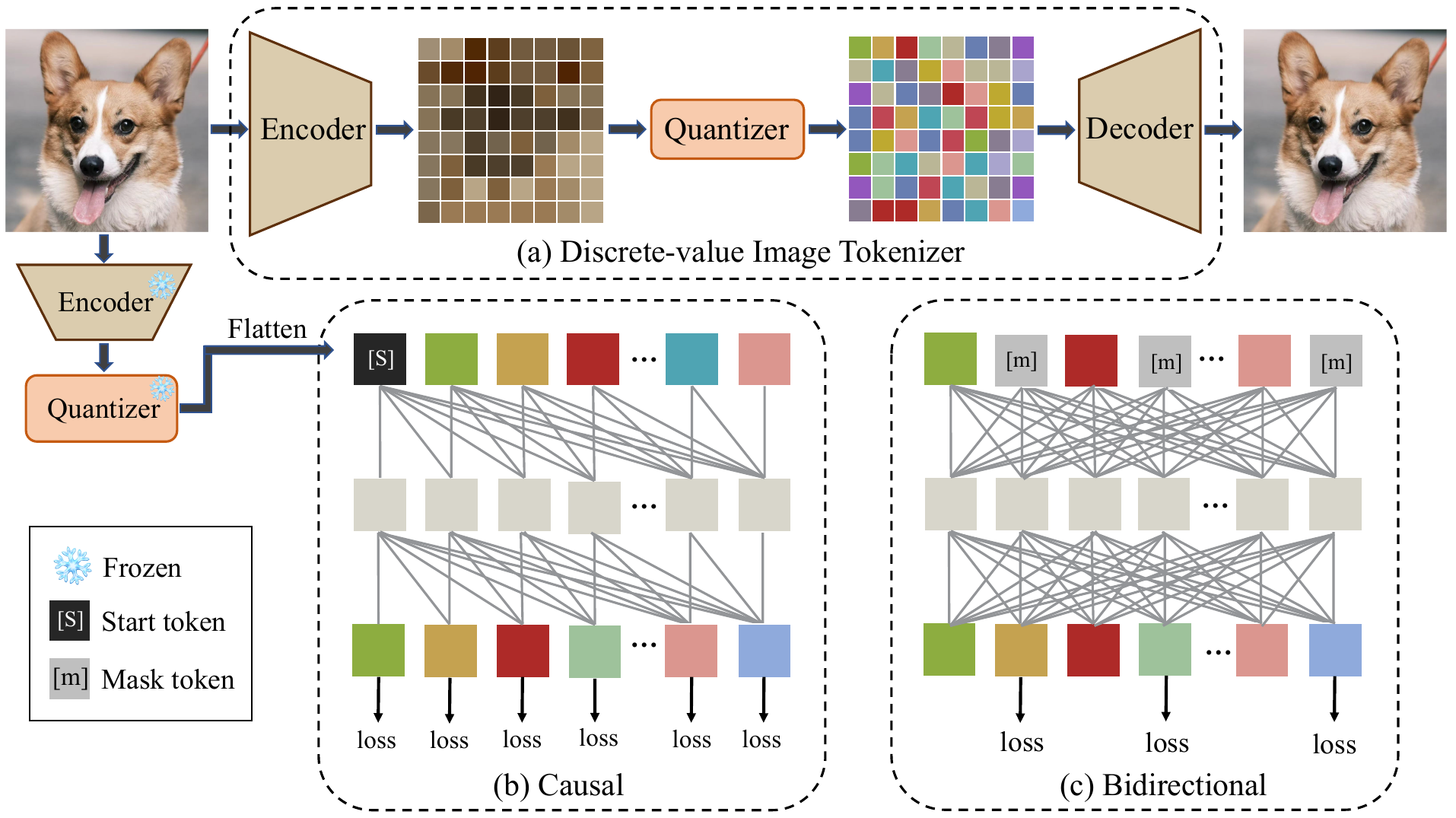}
    \caption{Illustration of (a) discrete-value image tokenizer (encoder and quantizer) and decoder via image reconstruction, (b) training the AR model through causal attention modeling and (c) training the MAR model through bidirectional attention modeling.}
    \label{fig:preliminaries}
\end{figure}
\begin{itemize}
    \item \textbf{Exploring the Optimal Design of Mask AutoRegressive Model.} We conduct a comprehensive study to evaluate the impact of different image tokenizers on image generation, aiming to identify the most effective tokenizer. Furthermore, we introduce an improved Bidirectional LLaMA model, which ultimately defines the optimal MAR architecture.

    \item \textbf{Scaling Mask AutoRegressive Model.} Building upon the optimal MAR generative architecture, we introduce a series of class-driven image generation models with parameter sizes ranging from 111M to 1.4B. Our largest model attains an FID score of 3.71 on the ImageNet 256$\times$256 benchmark with merely 8 inference steps. Furthermore, we develop a text-to-image model with 775M parameters, enabling the efficient synthesis of high-fidelity photorealistic images at arbitrary resolutions.

    \item \textbf{Extending MAR to Accelerate AR Generation.} To achieve both high efficiency and superior image quality, we introduce a fusion framework that integrates the AR and MAR paradigms during inference. By leveraging partially generated tokens from AR as initialization for MAR, our approach facilitates a seamless transition between these two paradigms, offering a flexible trade-off between computational efficiency and generation fidelity. Notably, our framework reduces generation time by over 70+\% while delivering results comparable to state-of-the-art AR models.

    \item \textbf{Real-Time Speech-To-Image Generation System.} We develop a real-time speech-to-image generation system, wherein speech inputs are first translated into English text using Whisper, a multilingual speech recognition model supporting over 90 languages. The translated text then serves as a prompt to guide the MAR model in generating corresponding images, which are subsequently returned to the user.
\end{itemize}

\section{Preliminaries}
\subsection{Discrete-Value Image Tokenizer}
To begin with, we revisit the discrete-value image tokenizer, which plays a crucial role in both AR and MAR image generation. The most commonly used tokenizer is the VQ-VAE~\citep{vq-vae}, an encoder-quantizer-decoder architecture, as shown in Figure~\ref{fig:preliminaries}\textcolor{red}{(a)}. This architecture employs a ConvNet for both the encoder and decoder, featuring a downsampling ratio $p$. The encoder projects the image pixels $x \in \mathbb{R}^{H\times W \times 3}$ to a feature map $f\in \mathbb{R}^{h \times w \times C}$, where $h=H/p$ and $w=W/p$. The core of this process lies in the quantizer, which includes a codebook $Z\in \mathbb{R}^{K\times C}$ with $K$ learnable vectors. Each vector $f^{(i, j)}$ in the feature map is mapped during quantization to the code index $q^{(i, j)}$ of its nearest vector $z^{k}$ in the codebook. Consequently, the image pixels $x \in \mathbb{R}^{H\times W \times 3}$ are quantized into $q \in \mathbb{Q}^{h \times w}$. During the decoding phase, the code index $q^{(i, j)}$ is remapped to the feature vector and the decoder converts these feature vectors back to the image pixels $\hat{x}$. There are various works~\citep{vavqe2,vqgan,yulanguage} that continue to explore the design of VQ-VAE for improving reconstruction quality. 

\begin{table*}[!t]
    \centering
    \setlength{\tabcolsep}{7pt}
    \renewcommand\arraystretch{1.1}
    % \vspace{-0.3cm}
    \caption{Reconstruction performance and codebook usage of different discrete-value image tokenizers. All tokenizers employ a downsampling ratio of 16 and are trained on the ImageNet training set at a training resolution of $256 \times 256$.}
    \resizebox{0.95\linewidth}{!}{
        \begin{tabular}{lcccccc}
        \toprule
        \multirow{2}{*}{\textbf{Method}} &\multirow{2}{*}{\textbf{Year}}  & \multirow{2}{*}{\textbf{Ratio}} & \textbf{Train} & \textbf{Codebook} & \multirow{2}{*}{\textbf{rFID}$\downarrow$} & \textbf{Codebook} \\
         &  & & \textbf{Resolution} & \textbf{Size} & & \textbf{Usage}$\uparrow$ \\
        \midrule
        MaskGIT-VQ~\citep{besnier2023pytorch} &2023 & $16 \times 16$  & $256 \times 256$ & 1024 & 10.79 & 44.3\% \\
        Chameleon-VQ~\citep{chameleon} &2024 & $16 \times 16$ & $256 \times 256$ & 8192 & 8.34 & 38.3\% \\
        LlamaGen-VQ~\citep{llamagen} &2024 & $16 \times 16$ & $256 \times 256$ & 16384 & 4.54 & 100\% \\
        Open-MAGVIT2-VQ~\citep{luo2024open-magvit2} &2024 & $16 \times 16$ &$256 \times 256$ & 262144 & 4.03 & 100\% \\ 
    \bottomrule
    \end{tabular}}
    \label{tab:rfid}
\end{table*}

\subsection{AutoRegressive Generative Models}
AutoRegressive models revolutionize the fields of language modeling~\citep{gpt3,gpt4,gemini,llama,llama3} and multimodal understanding~\citep{llava,video-llava,chameleon} using the unified \textbf{\textit{next-token prediction paradigm for all modalities with a casual transoformer}}, as illustrated in Figure~\ref{fig:preliminaries}\textcolor{red}{(b)}. 

This paradigm has been extended to the visual generation domain by early works such as DALL-E~\citep{dalle}, Cogview~\citep{cogview}, and Parti~\citep{parti}. These works leverage a two-stage approach where the image tokenizer first encodes continuous images into discrete tokens then the transformer decoder models the flattened one-dimensional sequences. During training, the casual transformer is trained to predict the categorical distribution $p_{\theta}(x_i\mid x_{\textless i};c)$ at each position conditioned on the additional information $c$, e.g., text prompts or class labels. During inference, image token sequences can be sampled in the same way as language generation and further decoded back to pixels. Despite this simple and unified paradigm for image synthesis, AR-based visual generative models have long been overlooked for a while, particularly after the exploding of diffusion models. One potential reason is their inferior generation quality restricted by the image tokenizer. Recently, LlamaGen~\citep{llamagen} improves the design of image tokenizer and leverages the Llama architecture for scaling, which enhance the performance of autoregressive models.

\subsection{Mask AutoRegressive Generative Models}
Different from the next token prediction paradigm typical of autoregressive generation, mask autoregressive models~\citep{chang2022maskgit,li2023mage,muse} leverage a \textbf{\textit{bidirectional transformer to simultaneously generate all visual tokens through a masked-prediction mechanism}}, as illustrated in Figure~\ref{fig:preliminaries}\textcolor{red}{(c)}. 

These models are trained using a proxy task similar to the mask prediction task employed in BERT~\citep{kenton2019bert}. In this setup, bidirectional attention allows all known tokens to see each other while also permitting all unknown tokens to view all known tokens, enhancing communication across tokens compared to causal attention. In contrast to causal attention, where training loss is computed on sequentially revealed tokens, MAR models compute the loss solely on the unknown tokens. At inference time, these models utilize a novel decoding method that synthesizes an image in a constant number of steps, typically between 8 and 15~\citep{chang2022maskgit}. Specifically, during each iteration, the model predicts all tokens in parallel, retaining only those predicted with high confidence. Less certain tokens are masked and re-predicted in subsequent iterations. This process repeats, progressively reducing the mask ratio until all tokens are accurately generated through several refinement iterations.

\begin{figure*}[t]
   \begin{picture}(0,155)
     \put(-18,6){\includegraphics[width=0.38\linewidth]{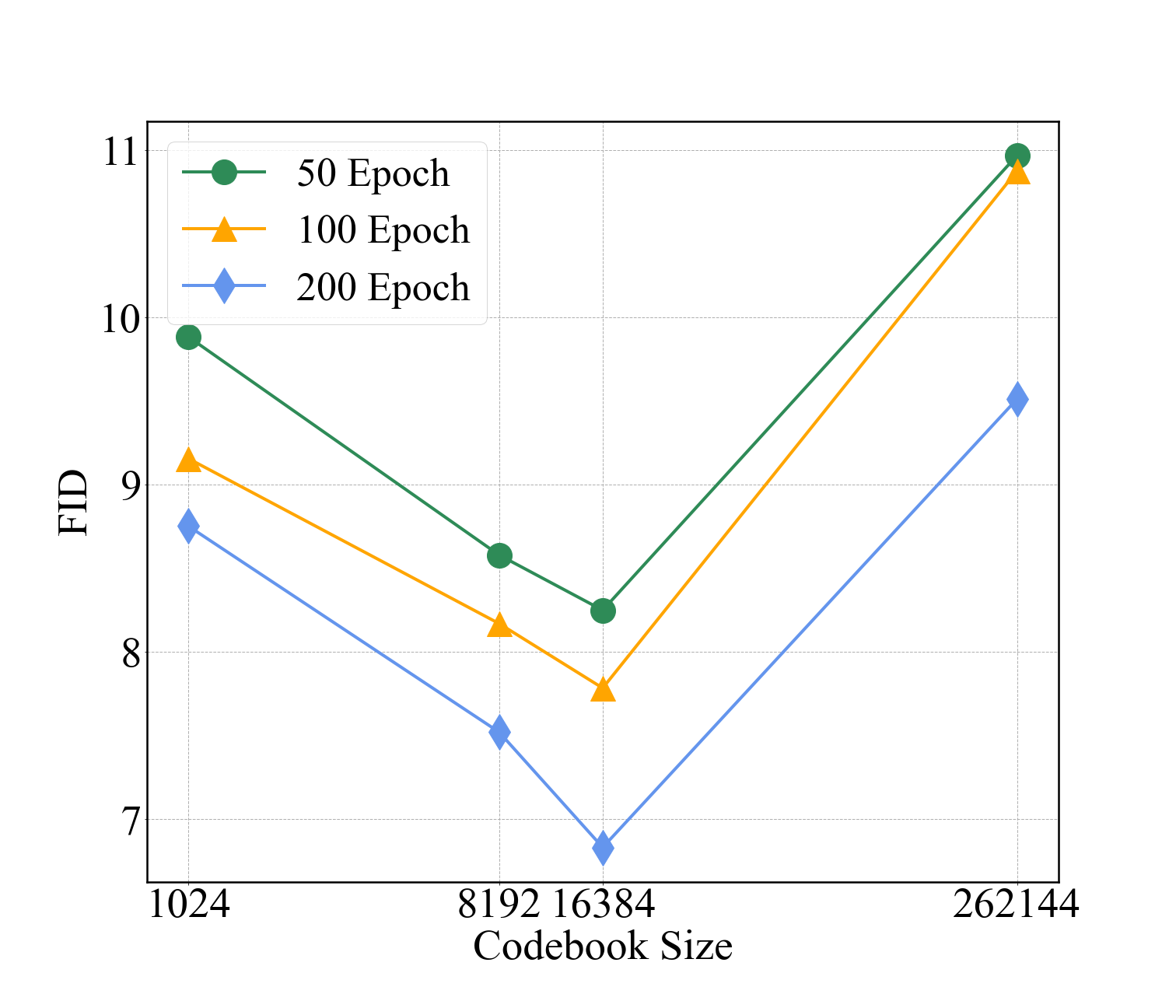}}
     \put(145,6){\includegraphics[width=0.38\linewidth]{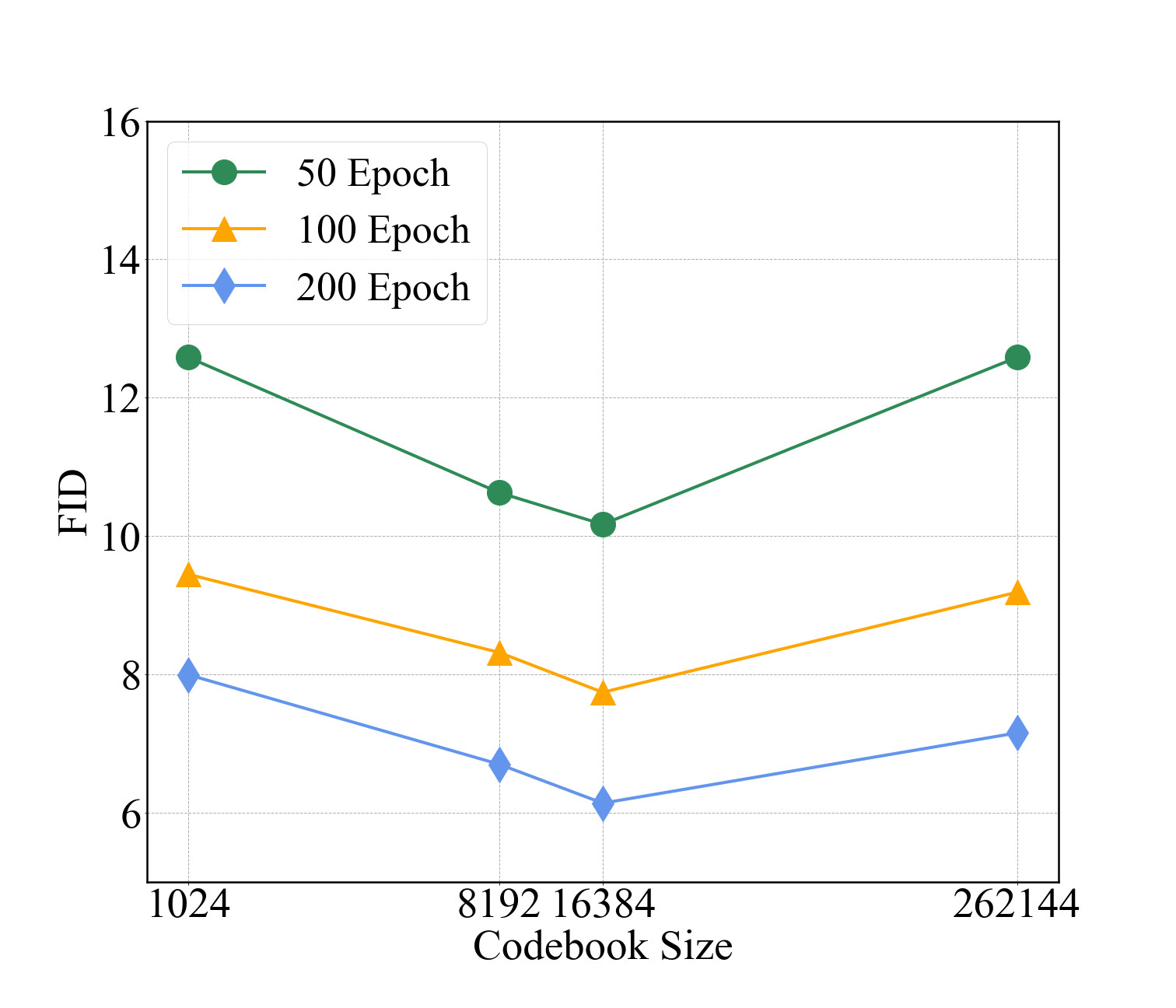}}
     \put(306,6){\includegraphics[width=0.38\linewidth]{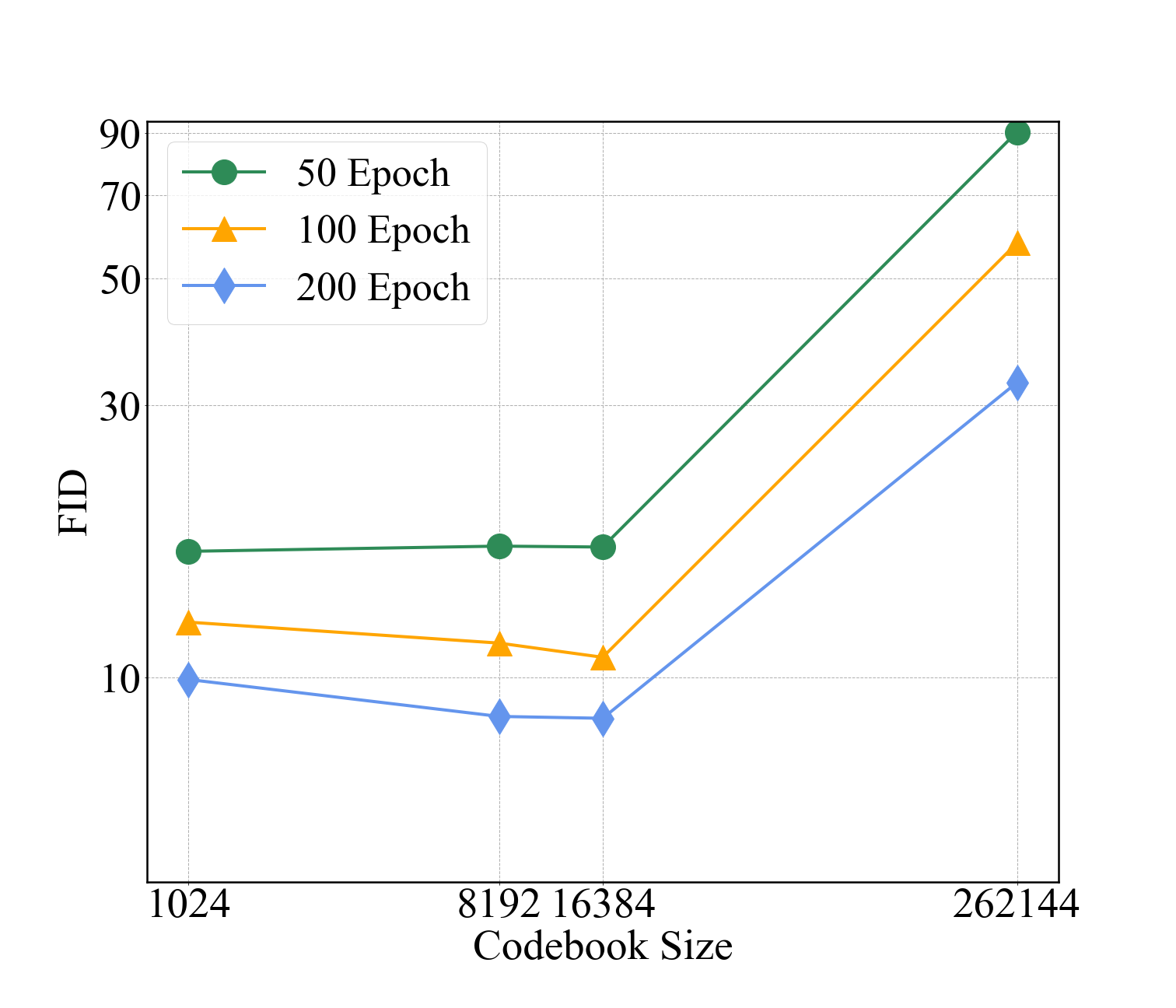}}
     \put(30,2){\small (a) \textit{\textcolor{teal}{Causal LLaMA (AR)}}}
     \put(170,2){\small (b) \textit{\textcolor{orange}{Bidirectional LLaMA (MAR)}}}
     \put(320,2){\small (c) \textit{\textcolor{red}{Bidirectional Transformer (MAR)}}}
   \end{picture}
    \caption{Visual Generation Evalutation. We show FID of class-driven ImageNet 256$\times$256 benchmark over training epochs.}
    \label{fig:vq_compare}
\end{figure*}
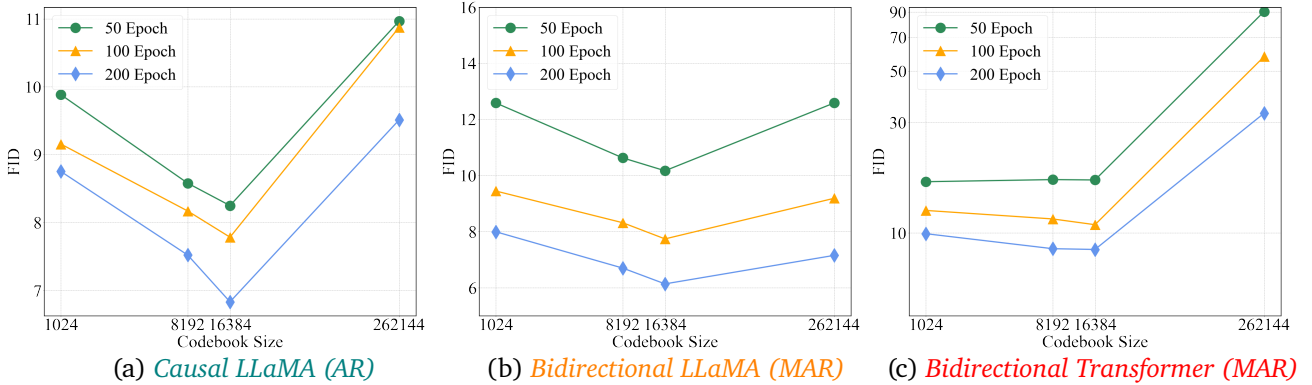
\section{Exploring the Optimal Design of Mask AutoRegressive Models}

\subsection{Rethinking Image Tokenizer For High-Quality Generation}
\label{sec: rethinking}
For high-quality image generation, the choice of a discrete-value image tokenizer is critical, as it determines the upper limit of the generation quality. Various tokenizers have been developed, and in this study, we evaluate four recent mainstream tokenizers, including MaskGIT-VQ~\citep{besnier2023pytorch}, Chameleon-VQ~\citep{chameleon}, LlamaGen-VQ~\citep{llamagen}, and Open-MAGVIT2-VQ~\citep{luo2024open-magvit2}, as shown in Table~\ref{tab:rfid}. All tokenizers use a downsampling ratio of 16$\times$16 and are trained on the ImageNet~\citep{deng2009imagenet}. We conduct detailed experiments using both AR and MAR frameworks to provide comprehensive insights into image tokenizers.

\subsubsection{Visual Reconstruction Evaluation}
\label{sec:vq_recon_eval}
Before assessing the impact of the image tokenizer on generation quality, we initially focus on evaluating the performance of each tokenizer itself. We analyze the reconstruction quality and codebook utilization using the ImageNet validation set, with results detailed in Table~\ref{tab:rfid}. From the results, we can find that as the codebook size increases, the relative rFID\footnote{The reconstruction  Fréchet inception distance, denoted as rFID~\citep{heusel2017gans}, is usually adopted to measure the quality of reconstructed images.} value decreases, indicating that the reconstruction image quality is gradually improving. However, when the codebook size reaches a certain level, the improvement in the reconstructed image quality is limited. For instance, despite Open-MAGVIT2-VQ expanding its codebook size 16$\times$ compared to LlamaGen-VQ, the rFID reduction is a mere 0.51, indicating that blindly increasing the codebook size for the sole purpose of improvement is not a wise choice in the design of a discrete-value image tokenizer. In terms of codebook utilization, LlamaGen-VQ and Open-MAGVIT2-VQ achieve 100\%, which is due to replacing traditional code assignment (\textit{i.e.}, pair-wise distance) with lookup-free quantization~\citep{luo2024open-magvit2}.

\subsubsection{Visual Generation Evaluation}
\label{sec:vq_generation_eval}
\vspace{0.1cm}  \noindent \textbf{Evaluation Setting.} To explore the impact of image tokenizers on image generation, we conduct detailed experiments using four tokenizers across two paradigms: AutoRegressive (AR) and Mask AutoRegressive (MAR) generation. For AR generation, we utilize the advanced LLaMA as the foundational architecture, consistent with LlamaGen~\citep{llamagen}. For MAR generation, following MaskGIT~\citep{chang2022maskgit}, we employ the Bidirectional Transformer model as the foundational architecture and also modify LLaMA to a bidirectional variant. To facilitate a fair comparison between these foundational models, both the Transformer and LLaMA architectures are configured under identical conditions ($\sim$ 100M parameters): 12 layers, 12 heads, and 768 dimensions. All experiments are conducted on the class-driven image generation ImageNet benchmark with 256$\times$256 resolution and trained for 200 epochs. During the evaluation, we generate 50,000 images across 1,000 classes, with 50 images per class. We employ FID and Inception Score (IS) as evaluation metrics, consistent with previous studies~\citep{li2024autoregressive}. 

\vspace{0.1cm}  \noindent \textbf{Evaluation Results.} The evaluation results are shown in Figure~\ref{fig:vq_compare}, and the detailed results are shown in Appendix~\ref{appendix:image_tokenizer}. Based on the results, LlamaGen-VQ emerges as the superior choice, delivering the best performance in both AR and MAR image generation tasks. Open-MAGVIT2-VQ, which performs best in the reconstruction evaluation, delivers unsatisfactory results in image generation quality, especially on the Transformer architecture of the MAR generation task, where it performs very poorly. As for Chameleon-VQ, although it is widely adopted in many previous image generation works~\citep{chameleon,liu2024lumina}, its performance does not match that of LlamaGen-VQ. Therefore, our exploration provides guidance for the selection of image tokenizers, suggesting that LlamaGen-VQ may be a better choice.
\begin{table*}[t]
\centering
\caption{Model sizes and architecture configurations of MaskGIL models. The configurations are following previous works~\citep{llamagen,llama,open_Llama_3b}. We replace causal attention to bidirectional attention for MAR modeling. We add QK-Norm and Post-Norm for stable training.
}
\resizebox{0.88\linewidth}{!}{
\begin{tabular}{@{}lcccccc@{}}
\toprule
\bf Model & \bf Parameters & \bf Layers & \bf Hidden Size & \bf Heads &\bf QK-Norm & \bf Post-Norm \\
\midrule
MaskGIL-B & 111M & 12 & 768 &  12 &\ding{56} &\ding{56}\\
MaskGIL-L & 343M & 24 & 1024 &  16 &\ding{56} &\ding{56}\\
MaskGIL-XL & 775M & 36 & 1280 &  20 &\ding{56} &\ding{56}\\
MaskGIL-XXL & 1.4B & 48 & 1536 &  24 &\ding{52} &\ding{52}\\
\bottomrule
\end{tabular}
}
\label{tab:model_scaling}
\end{table*}

\vspace{0.1cm}  \noindent \textbf{Discussion.} Our evaluation focuses on generative models with $\sim$ 100M parameters, providing a valuable benchmark. While we acknowledge the possibility that scaling model parameters, Open-MAGVIT2-VQ, with its 262,144 codebook size, could potentially yield better performance. However, its high training complexity, large parameters for the classification head, and significant GPU resource demands make it impractical for many real-world applications. Furthermore, all evaluations are only conducted on class-driven generation tasks, which may introduce a slight bias to our experimental results, but this limitation has minimal impact on the overall conclusions.

\subsection{Optimized Mask Autoregressive Model Architecture}
In Section \textcolor{red}{3.1}, we investigate the influence of image tokenizers on generation quality and identify LlamaGen-VQ as the most effective tokenizer. To further optimize the MAR architecture, we introduce a Bidirectional LLaMA architecture, inspired by the state-of-the-art AR-based model~\citep{llamagen}, referred to as \textbf{Mask}ed \textbf{G}enerative \textbf{I}mage \textbf{L}LaMA (\textbf{MaskGIL}). Specifically, we modify causal attention with bidirectional attention and integrate 2D RoPE into every layer of the LLaMA architecture. Moreover, we omit the use of the AdaLN technique~\citep{peebles2023scalable} to preserve alignment with the standard LLM architecture. A comparison between our \textit{\textcolor{orange}{Bidirectional LLaMA}} and the \textit{\textcolor{red}{Bidirectional Transformer}} used in the previous MAR generation, as depicted in Figure~\ref{fig:vq_compare}, demonstrates that the Bidirectional LLaMA consistently achieves superior generation quality.

\section{Resurrect Mask AutoRegressive Model}
\label{sec:4_scaling}
Numerous studies have extensively explored AR generation, with prominent works such as LlamaGen~\citep{llamagen}, which successfully scaled AR models to 3B parameters. In contrast, the scaling of MAR models remains relatively underexplored, primarily due to their performance gap compared to AR models. However, our optimized MaskGIL architecture effectively bridges this gap, resurrecting the MAR generation paradigm and unlocking its potential for broader applications.

\subsection{Model Scaling}
\vspace{0.1cm}  \noindent \textbf{Scaling Mask AutoRegressive Model.} Previous research on MAR image generation~\citep{chang2022maskgit, li2023mage} has primarily focused on models with parameter sizes ranging between 200M and 300M, with limited exploration into scaling to larger model sizes. In this study, we examine the relationship between the generation performance of MAR models and model scaling by increasing the parameter size of our MaskGIL from 111M to 1.4B on the class-driven ImageNet generation benchmark. The detailed configurations of our MaskGIL models, with varying parameter sizes, are provided in Table~\ref{tab:model_scaling}. Additionally, we develop text-driven MaskGIL models with 775M parameters, capable of generating images at any resolution according to the text descriptions.

\vspace{0.1cm}  \noindent \textbf{Training Stability.} 
Scaling MaskGIL models beyond 1.4B parameters presents significant challenges in maintaining stable training, as instabilities often arise during the late stages of training. This observation is consistent with findings from several prior studies~\citep{dehghani2023scaling, chameleon, gao2024lumina, zhuo2024lumina}. The fundamental reason stems from the standard LLaMA architecture for visual modeling, which exhibits complex divergences due to slow norm growth during the mid-to-late stages of training. To address this problem, we adopt query-key normalization (QK-Norm)~\citep{chameleon, dehghani2023scaling} and Post-Norm~\citep{zhuo2024lumina}. QK-Norm involves applying layer normalization to the query and key vectors within the attention mechanism, while Post-Norm applies layer normalization to the outputs of both the attention and MLP layers. Together, these techniques effectively mitigate uncontrollable norm growth caused by unnormalized pathways, thereby stabilizing the training process and enabling the successful scaling of MaskGIL models.

\subsection{Robust and Efficient Inference Strategy}
\vspace{0.1cm} 
\noindent \textbf{Classifier-Free Guidance.} Classifier-Free Guidance (CFG)~\citep{ho2021classifier,sanchezstay} was initially introduced to enhance the quality and text alignment of generated samples in text-to-image diffusion models. We incorporate this technique into our MaskGIL models. During training, the conditional input is randomly dropped and replaced with a null unconditional embedding~\citep{liu2024lumina,zhuo2024lumina,yi2024towards,zhuo2025reflection,qin2025lumina}. During inference, for each image token, CFG modifies the logits $\ell_{cfg}$, which are defined as $\ell_{cfg}=\ell_{u}+s(\ell_{c}-\ell_{u})$. Here, $\ell_{c}$ represents the conditional logits, $\ell_{u}$ denotes the unconditional logits, and $s$ is the scale factor for the classifier-free guidance. As illustrated in Figure \ref{fig:ablation}\textcolor{red}{(b)}, CFG substantially influences the performance of MaskGIL models.

\vspace{0.1cm}  
\noindent \textbf{Iterative Decoding.} Despite the bidirectional architecture of MaskGIL enabling one-step image generation, the generated images might lack sufficient quality. To address this, we adopt the iterative decoding strategy proposed in~\citep{chang2022maskgit}, which facilitates image generation over $T$ steps, typically ranging from 8 to 16. \textbf{\textit{At each iteration, the model predicts all tokens simultaneously but retains only the most confident predictions.}} Tokens with lower confidence are masked and re-predicted in subsequent iterations, with the mask ratio progressively decreasing until all tokens are generated within the designated $T$ iterations.

To generate an image, we begin with a blank canvas where all tokens are initially masked, denoted as $Y_{M}^{(0)}$. At iteration $t$, the iterative decoding algorithm proceeds as follows:
\begin{itemize}
    \item \textbf{Step 1: Predict.} Given the masked tokens $Y_M^{(t)}$ at the current iteration, the model predicts the probabilities, denoted as $p^{(t)} \in \mathbb{R}^{N \times K}$, for all masked locations in parallel.

    \item \textbf{Step 2: Sample.} For each masked location $i$, a token $y_i^{(t)}$ is sampled based on its prediction probabilities $p_i^{(t)} \in \mathbb{R}^K$ over all possible tokens in the codebook. The corresponding prediction score is used as a confidence score, representing the model's certainty in the prediction. Confidence scores for unmasked positions in $Y_M^{(t)}$ are set to 1.0.

     \item \textbf{Step 3: Mask Schedule.} The number of tokens to be masked is determined by the mask scheduling function $\gamma$, computed as $n = \lceil \gamma(\frac{t}{T}) N \rceil$, where $N$ is the input length and $T$ is the total number of iterations. In this work, we design five types of mask schedules: Root Schedule, Linear Schedule, Cosine Schedule, Square Schedule, and Arccos Schedule. For detailed descriptions of these schedules, please refer to Appendix~\ref{sec:sch_appendix}.

     \item \textbf{Step 4: Mask.} The updated masked tokens $Y_M^{(t+1)}$ are obtained by masking $n$ tokens in $Y_M^{(t)}$. The mask $M^{(t+1)}$ for iteration $t+1$ is calculated as:
        \begin{equation*}
         m_{i}^{(t+1)} = 
        \begin{cases}
            1,  & \text{if $c_i < {\text{sorted}}_j(c_j)[n]$.}\\
            0,  & \text{otherwise.}\vspace{-2mm}
        \end{cases},
        \end{equation*}
    where $c_i$ is the confidence score for the $i$-th token.
\end{itemize}

\subsection{Class-Driven Image Generation} 
\vspace{0.1cm} 
\noindent \textbf{Training and Evaluation Setup.} The class embedding is derived from a set of learnable embeddings~\citep{li2024scalable,vqgan} and serves as the prefilling token embedding. From this initial token embedding, the model generates all image tokens. We conduct class-driven generation experiments on the ImageNet dataset. All models are trained for token unmasking using cross-entropy loss with a label smoothing factor of 0.1. The optimizer used is AdamW with a learning rate of $1e^{-4}$, betas=(0.9, 0.96), and a weight decay of $1e^{-5}$. For classifier-free guidance, the dropout rate for the class condition embedding is set to 0.1. All models are trained with a batch size of 256. In addition to evaluating performance with the FID and IS metrics, we also report Precision and Recall~\citep{kynkaanniemi2019improved} to provide a more comprehensive evaluation.
\begin{table*}[!t]
\centering
\renewcommand{\arraystretch}{1.25}  % 调整行高
\setlength{\tabcolsep}{4.5pt} % 调整列间距
\caption{Model comparisons on class-conditional ImageNet 256$\times$256 benchmark. ``Step'' indicates the number of inference steps. Metrics include FID, IS, Precision and Recall.
``$\downarrow$'' or ``$\uparrow$'' indicate lower or higher values are better. $\dagger$ indicates that the result comes from the original paper.
}
\resizebox{0.9\linewidth}{!}{%
\begin{tabular}{c|lcc|cccc}
\toprule

\bf Type &\bf Model &\bf \#Param. &\bf Steps$\downarrow$ &\bf FID$\downarrow$ &\bf IS$\uparrow$ & \bf Precision$\uparrow$ &\bf Recall$\uparrow$  \\
\midrule

\multirow{5}{*}{AR} &RQTransfomer$^{\dagger}$~\citep{lee2022autoregressive} &3.8B &256 &7.55 &134.00 & $-$ & $-$ \\

& LlamaGen-B$^{\dagger}$~\citep{llamagen} & 111M &256 & 5.46 & 193.61 & 0.83 & 0.45\\
 & LlamaGen-L$^{\dagger}$~\citep{llamagen} & 343M &256 & 3.07 & 256.06 & 0.83 & 0.52 \\
 & LlamaGen-XL$^{\dagger}$~\citep{llamagen} & 775M &256 & 2.62 & 244.08 & 0.80 & 0.57 \\
 & LlamaGen-XXL$^{\dagger}$~\citep{llamagen} & 1.4B &256 & 2.34 & 253.90 & 0.80 & 0.59 \\
\midrule
 \multirow{2}{*}{MAR} & MaskGIT$^{\dagger}$~\citep{chang2022maskgit}  & 227M   &8 & 6.18  & 182.1        & 0.80 & 0.51  \\
 & MAGE$^{\dagger}$~\citep{li2023mage} & 230M &20   &6.93   & 195.8       & $-$ & $-$ \\

\midrule
\multirow{4}{*}{MAR} 
& MaskGIL-B (CFG=2.0)   & 111M     &8  &5.64  &229.96  &0.83  &0.48    \\
& MaskGIL-L (CFG=2.0)   & 343M     &8  &4.01  &281.11  &0.84  &0.51    \\
& MaskGIL-XL (CFG=2.5)  & 775M    &8  &3.90  &296.25  &0.87  &0.49    \\
& MaskGIL-XXL (CFG=2.5) & 1.4B    &8  &3.71  &303.47  &0.88  &0.52    \\
\bottomrule
\end{tabular}
\label{tab:nar_scaling_c2i}
}
\end{table*}

\begin{figure*}[ht]
   \begin{picture}(0,150)
     \put(-18,8){\includegraphics[width=0.37\linewidth]{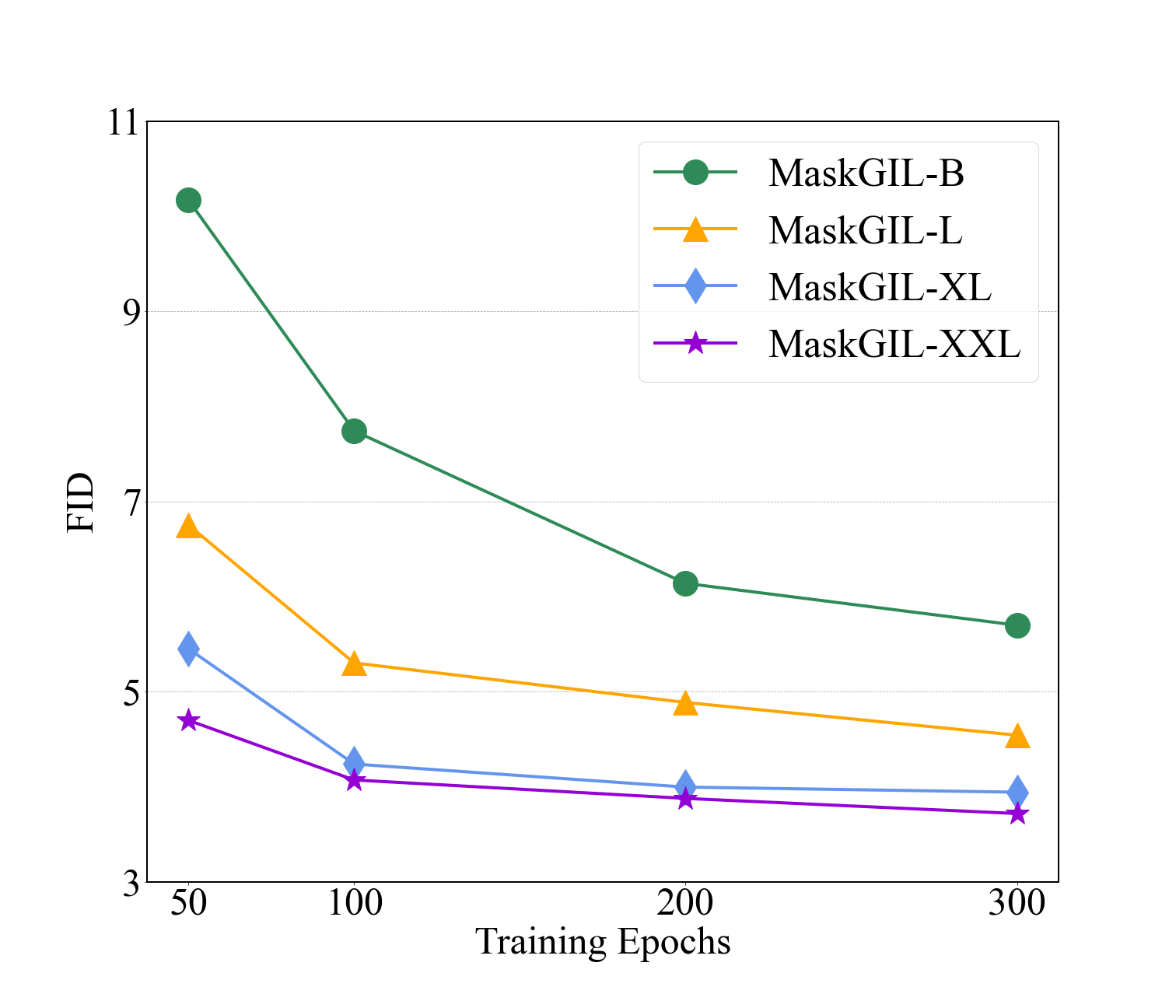}}
     \put(138,8){\includegraphics[width=0.37\linewidth]{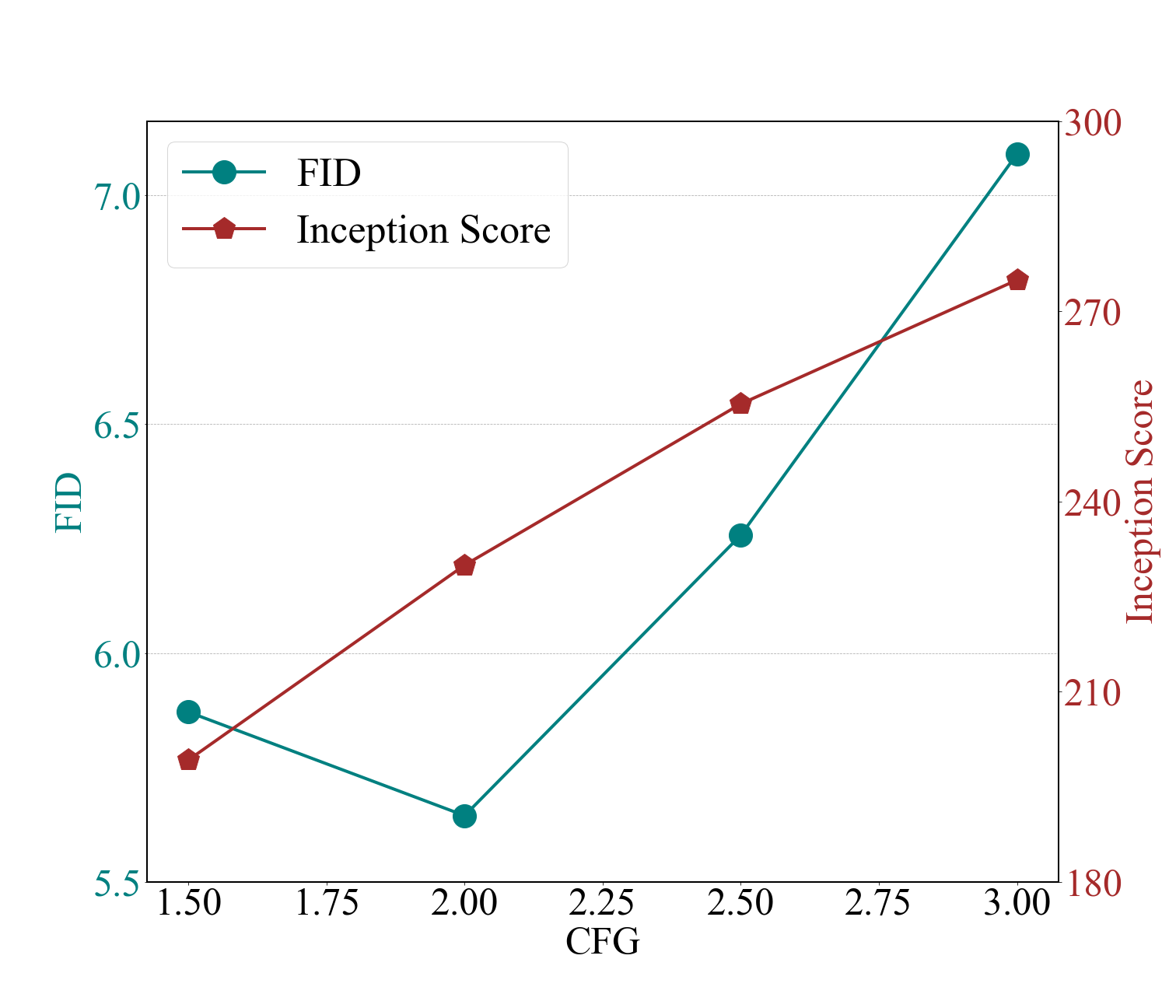}}
     \put(313,8){\includegraphics[width=0.37\linewidth]{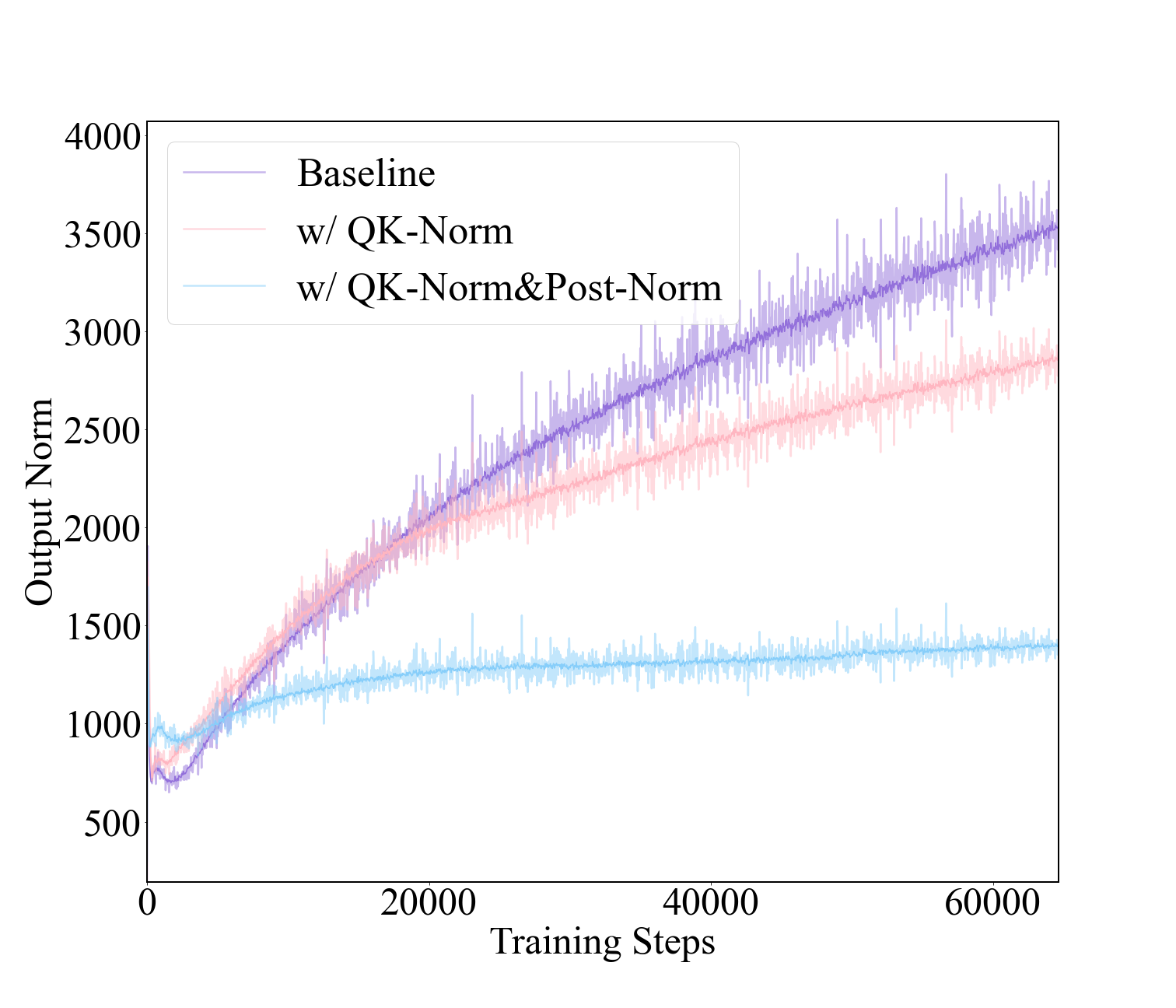}}
     \put(42,3){\small (a) Model Size}
     \put(167,3){\small (b) Classifier-Free Guidance}
     \put(340,3){\small (c) QK-Norm and Post-Norm}
   \end{picture}
    \caption{The Effect of Model Size, CFG, QK-Norm, and Post-Norm. We show the FID scores on the ImageNet benchmark across different model sizes and CFG configurations. Scaling the model size consistently improves FID scores throughout the training process. The impact of CFG is also notable. To monitor training stability, we plot the model’s output norm.}
    \label{fig:ablation}
    \vspace{-0.3cm}
\end{figure*}
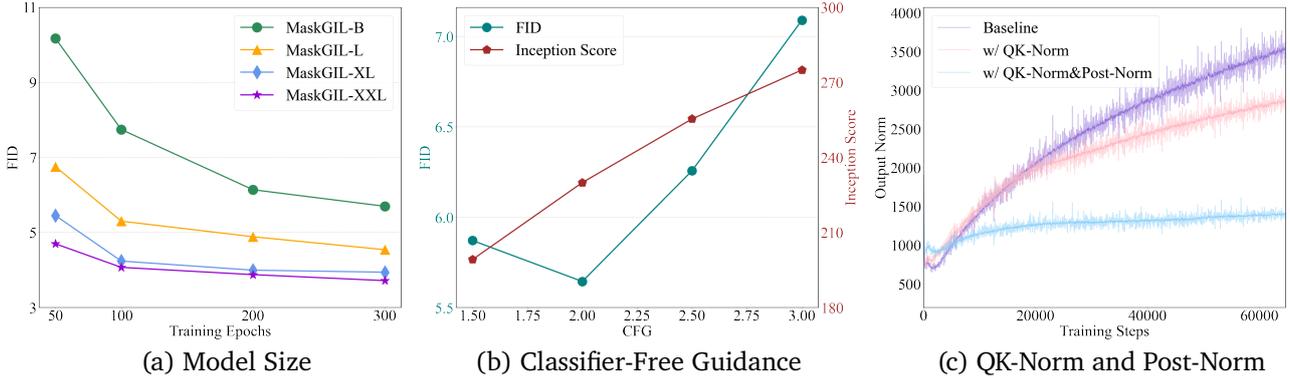
\vspace{0.1cm}  
\noindent \textbf{Comparisons with Other Class-Driven Generation Models.}  
In Table~\ref{tab:nar_scaling_c2i}, we compare our model against popular AR and MAR image generation models, including RQTransformer~\citep{lee2022autoregressive}, LlamaGen~\citep{llamagen}, MaskGIT~\citep{chang2022maskgit}, and MAGE~\citep{li2023mage}. Our MaskGIL models achieve competitive performance across all metrics, including FID, IS, Precision, and Recall. \textbf{\textit{Importantly, MaskGIL requires only 8 inference steps, underscoring its efficiency in image generation.}} Further details on inference steps are provided in Appendix~\ref{sec:decoding_steps}. While the FID and Recall scores of MaskGIL are slightly lower than those of LlamaGen, it achieves higher IS and Precision scores. This result indicates a trade-off between diversity and visual quality in the generated images, which could potentially be mitigated by exploring more advanced sampling strategies.

\vspace{0.1cm}  
\noindent \textbf{Effective of Model Size.} We train our MaskGIL models across four model sizes (B-111M, L-343M, XL-775M, XXL-1.4B) and evaluate their performance in Table~\ref{tab:nar_scaling_c2i}. Figure~\ref{fig:ablation}\textcolor{red}{(a)} depicts the changes in FID as both model size and training epochs increase. Significant improvements in FID are observed when scaling the model from MaskGIL-B to MaskGIL-XL. However, further scaling to 1.4B results in only marginal gains. A plausible explanation for this phenomenon is that the ImageNet dataset size may constrain the performance improvements achievable through model scaling~\citep{llamagen}.

\begin{figure*}[t]
    \centering
    \includegraphics[width=1.0\linewidth]{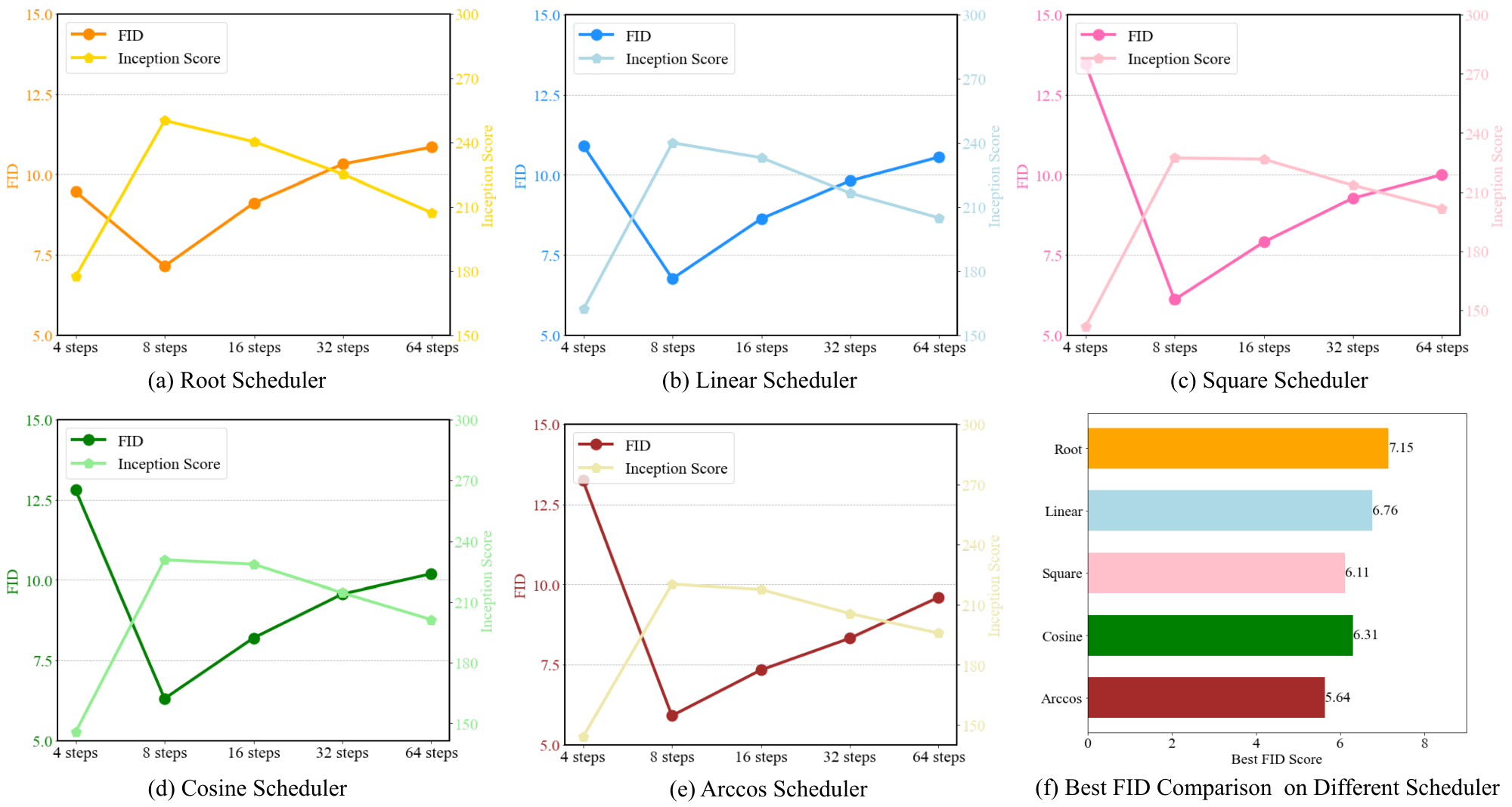}
    \caption{Performance Comparison of Different Decoding Steps and Different Schedulers. This experiment shows the checkpoint results of MaskGIL-B at 400 epochs, with CFG$=2.0$.}
    \label{fig:DecodingSteps}
\end{figure*}
\vspace{0.1cm}  \noindent \textbf{Effective of Classifier-Free Guidance.} 
Figure~\ref{fig:ablation}\textcolor{red}{(b)} presents the FID and IS scores of MaskGIL-L under various classifier-free guidance (CFG) settings. MaskGIL-L achieves its best FID at CFG = 2.0, and increasing CFG beyond this point results in a deterioration in FID, which aligns with previous findings~\citep{dhariwal2021diffusion, llamagen}. This demonstrates that CFG plays a crucial role in influencing the generation performance of our MaskGIL models. For further results on different model sizes, please refer to Appendix~\ref{sec:appendix_scaling}.

\vspace{0.1cm}  \noindent \textbf{Effective of QK-Norm and Post-Norm.} In Figure~\ref{fig:ablation}\textcolor{red}{(c)}, we show the output norm of the MaskGIL-XXL (1.4B) with and without QK-Norm/Post-Norm. Without QK-Norm and Post-Norm, the output norm grows uncontrollably, leading to non-convergence and the occurrence of ``nan'' loss values during the training. Normalizing the query and key embeddings before computing the attention matrix helps to avoid attention collapse but merely mitigates this norm growth. Ultimately, the inclusion of both QK-Norm and Post-Norm ensures the output remains stable, preventing dramatic increases and resulting in more stable training for scaling.

\vspace{0.1cm} \noindent \textbf{Ablation Study on Different Iterative Decoding Schedules.} We evaluate five mask schedulers: Root, Linear, Square, Cosine, and Arccos (details in Appendix~\ref{sec:sch_appendix}). We conduct a comparative analysis of these schedulers on the ImageNet benchmark, as presented in Figure~\ref{fig:DecodingSteps}\textcolor{red}{(f)}. The results indicate that among all schedulers, the arccos scheduler achieves the best FID, while the root scheduler attains the best IS. Overall, we recommend using the arccos scheduler. 

\begin{table*}[!t]
\centering
\renewcommand{\arraystretch}{1.25}  % 调整行高
\setlength{\tabcolsep}{6pt} % 调整列间距
\caption{Performance comparison across different T2I models on GenEval benchmark.}
\resizebox{1.0\linewidth}{!}{%
    \begin{tabular}{lcccccccc}
    \toprule
    
    \textbf{Methods} &\textbf{\# Params} &\textbf{Single Obj.} &\textbf{Two Obj.} &\textbf{Counting} &\textbf{Colors} &\textbf{Position} &\textbf{Color Attri.} &\textbf{Overall$\uparrow$}\\ 
    \midrule
    
    \multicolumn{9}{l}{\textbf{Diffusion Models}} \\ \hline
    SDv1.5~\citep{rombach2022high}      & 0.9B &0.97 &0.38 &0.35 &0.76 &0.04 &0.06 &0.43\\ 
    % Lumina-Next~\citep{zhuo2024lumina} & 1.7B  \\ 
    SDv2.1~\citep{rombach2022high}     & 0.9B &0.98 &0.51 &0.44 &0.85 &0.07 &0.17 &0.50\\ 
    PixArt-$\alpha$~\citep{chen2023pixart} & 0.6B &0.98 &0.50 &0.44 &0.80 &0.08 &0.07 &0.48\\ 
    % SDXL~\citep{podell2023sdxl}  & 2.6B & 0.74  & 0.39  & 0.23  & 0.55  & 82.43  & 86.76  & 80.91  & 74.65  &0.63 &0.54 &0.56     \\ 
    \midrule
    
    \multicolumn{9}{l}{\textbf{AutoRegressive Models}} \\ \hline
    LlamaGen~\citep{llamagen}    & 0.8B &0.71 &0.34 &0.21 &0.58 &0.07 &0.04 &0.32\\ 
    
    Chameleon~\citep{chameleon}   & 7B &- &- &- &- &- &- &0.39  \\ 
    
    Lumina-mGPT~\citep{liu2024lumina} & 7B &0.98 &0.77 &0.27 &0.82 &0.17 &0.32 &0.56\\
    
    % \rowcolor{green!10} 
    % \textbf{Lumina-mGPT 2.0} & 2.8B &\\
    \textbf{MaskGIL (Ours)} & 0.7B &0.98 &0.52 &0.24 &0.69 &0.24 &0.29 &0.49 \\
    
    \bottomrule
    \end{tabular}%
    \label{tab:t2i_performance}
}
\end{table*}

\begin{figure*}[!t]
    \centering
    \includegraphics[width=0.98\linewidth]{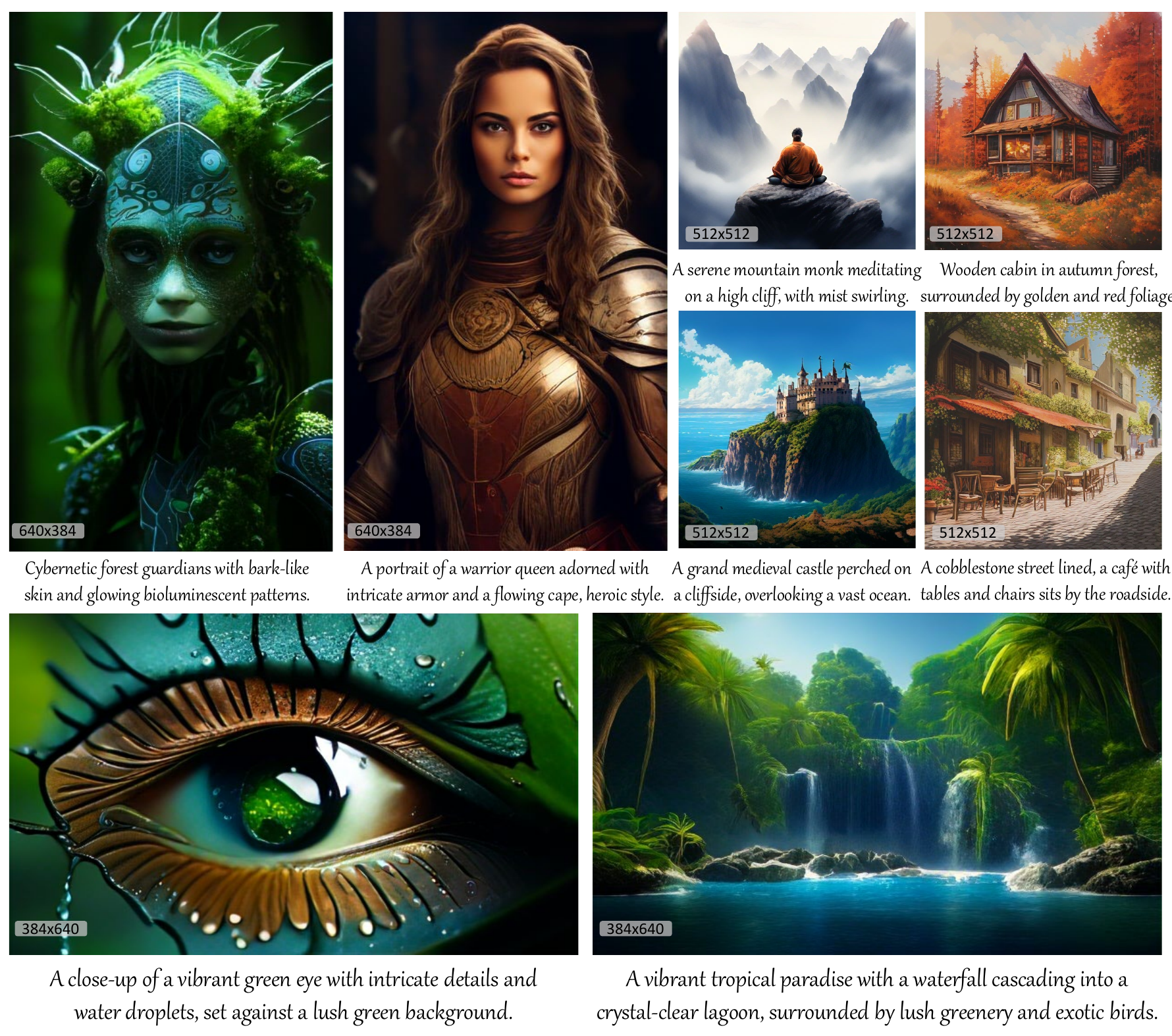}
    \caption{Visualization of text-driven image generation. The prompts are generated by GPT-4. MaskGIL can generate images at any resolution while preserving consistency with the text.}
    \label{fig:t2i_case}
\end{figure*}

\vspace{0.1cm}  \noindent \textbf{Ablation Study on Decoding Steps.} In Table~\ref{tab:nar_scaling_c2i}, all experimental results are based on an 8-step decoding process. To investigate how the number of decoding steps affects generation quality, we conduct an ablation study on the class-driven ImageNet benchmark, with findings illustrated in Figure~\ref{fig:DecodingSteps}\textcolor{red}{(a)-(e)}. The results reveal a consistent trend across all mask schedulers: increasing the number of decoding steps does not always lead to improved generation quality. Notably, the 8-step decoding process achieves the best results across all schedulers.

\begin{figure*}[!t]
    \centering    \includegraphics[width=1.0\linewidth]{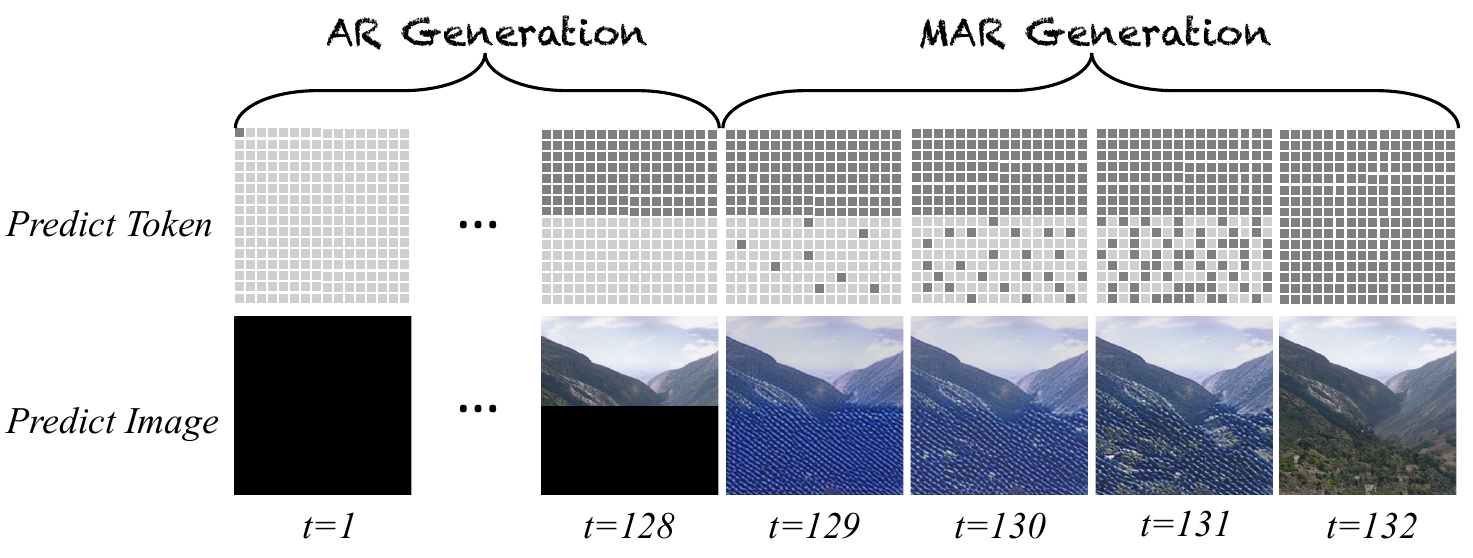}
    \caption{Illustration of our unified framework. This framework initially employs AR to generate a subset of image tokens. Subsequently, MAR is deployed to complete the generation of the remaining tokens. In this example, only 132 inference steps are needed, of which AR takes 128 steps.}
    \label{fig:framework}
\end{figure*}
\subsection{Text-Driven Image Generation}
\vspace{0.1cm}  \noindent \textbf{Image and Text Tokenizer.} While we evaluate the effectiveness of the LlamaGen-VQ tokenizer for class-driven generation, its application in text-driven generation remains unexplored due to its high resource consumption. Therefore, for text-driven generation, we adopt the Chameleon-VQ tokenizer, which is consistent with the advanced AR model. This choice also facilitates the exploration of AR and MAR integration, as discussed in Section~\ref{sec:unified}. To incorporate text prompts into our MaskGIL model, we utilize Gemma-2B~\citep{team2024gemma} as the text encoder. Gemma-2B offers performance comparable to large-scale LLMs while maintaining high efficiency. The extracted text features are subsequently processed through an additional MLP and used as prefilling token embeddings within MaskGIL.

\vspace{0.1cm} \noindent \textbf{Training and Evaluation Setup.} The training dataset consists of 2M high-aesthetic images with prompts generated by a mixture of captioners. We employ a two-stage progressive training pipeline, initially training on $256\times 256$ resolution before transitioning to $512\times 512$. To enable image generation with arbitrary aspect ratios, we adopt a multi-resolution training strategy by defining a series of size buckets and resizing each image to its nearest bucketed shape. All other training hyperparameters follow the class-driven settings. We employ standard GenEval~\citep{ghosh2024geneval} benchmark to evaluate MaskGIL. These metrics primarily assess text-image alignment.

\vspace{0.1cm}  \noindent \textbf{Quantitative Performance.} As shown in Table~\ref{tab:t2i_performance}, MaskGIL achieves competitive performance compared to existing AR and Diffusion models, obtaining an overall score of 0.49 on the GenEval benchmark. It surpasses LlamaGen and Chameleon but still lags behind the state-of-the-art model Lumina-mGPT. The primary factors contributing to this gap between MaskGIL and Lumina-mGPT are: 1) \textbf{Limited Training Data:} MaskGIL was trained on a significantly smaller dataset. 2) \textbf{Model Size:} MaskGIL contains only 0.7B parameters, whereas Lumina-mGPT has 7B parameters. If these two challenges are addressed—by scaling up training data and increasing model size—MaskGIL's capability in text-driven image generation is expected to improve significantly.

\vspace{0.1cm} \noindent \textbf{Qualitative Performance.} In Figure~\ref{fig:t2i_case}, we use prompts randomly generated by GPT-4o to create images at various resolutions. MaskGIL demonstrates strong text-to-image generation capabilities, producing high-quality images across diverse styles and subjects while maintaining excellent text alignment. In future work, we plan to extend support for higher resolution generation, such as $1024 \times 1024$.

\begin{figure*}[!t]
    \centering  \includegraphics[width=1.0\linewidth]{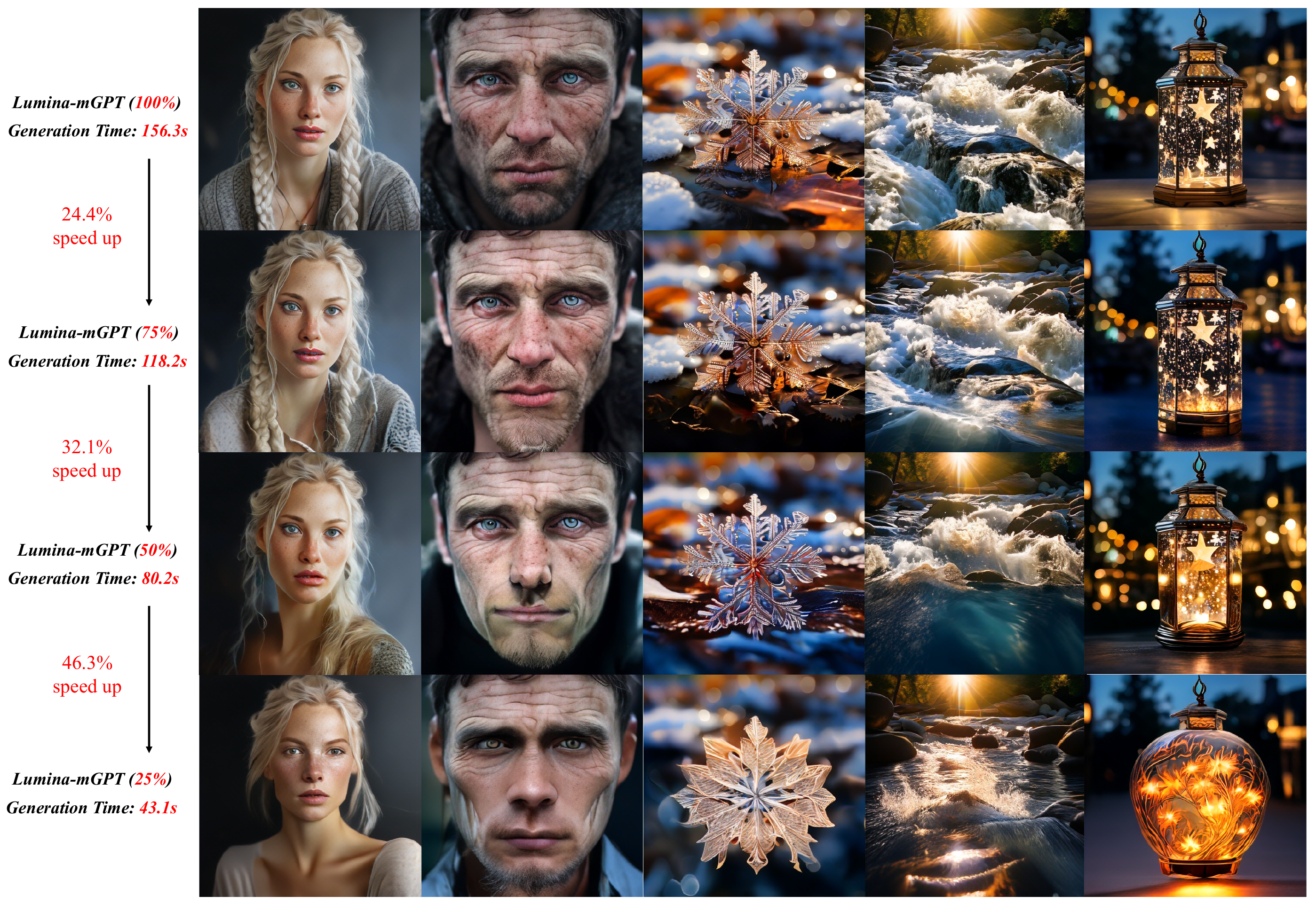}
    \caption{Generated images and time consumption at different ratios of inference steps of Lumina-mGPT. Our unified framework reduces generation time while ensuring quality.}
    \label{fig:t2i_case2}
\end{figure*}

\section{Application of MaskGIL}
\subsection{Extending MaskGIL to Accelerate AR Generation} 
\label{sec:unified}
From the exploration of AR and MAR generation paradigms above, it is evident that AR models excel in producing high-quality images, while MAR models offer the advantage of fast generation speed. To achieve both rapid and high-fidelity image generation, we propose a unified inference framework that integrates AR models with our MaskGIL, as shown in Figure~\ref{fig:framework}. We validate this framework on both class-driven and text-driven image generation tasks.

\vspace{0.1cm} \noindent \textbf{Unified Inference Framework for Integrating AR and MAR.} In this work, we extensively explore and optimize the generative capabilities of MAR models. However, despite these advancements, a performance gap remains when compared to AR models, as demonstrated in Table~\ref{tab:nar_scaling_c2i} and Table~\ref{tab:t2i_performance}. Drawing inspiration from recent research~\citep{li2024autoregressive}, which suggests that AR and MAR paradigms can be unified into a single generative framework, we propose a novel inference strategy that leverages the strengths of both approaches. As illustrated in Figure~\ref{fig:framework}, the proposed framework begins by employing AR models to generate a subset of image tokens, which serve as image/token prompts. These prompts are then used as input by MAR models to complete the generation of the remaining tokens. This hybrid approach enables efficient image generation while preserving high visual fidelity, effectively bridging the gap between speed and quality.

% \begin{table}[!t]
% \centering
% \setlength{\tabcolsep}{3pt}
% \renewcommand\arraystretch{1.4}
% \caption{Experimental Results of Sampling Steps and FID on ImageNet. ``AR Ratio'' refers to the proportion of sampling steps performed by the AR model relative to the initial total of 256 steps.}
% \resizebox{1.0\linewidth}{!}{%
% \begin{tabular}{l|ccc}
% \toprule
% \bf Method &\bf AR Ratio &\bf Sample Steps &\bf FID$\downarrow$  \\
% \midrule

% LlamaGen-L & 100\%  &256 &3.80 \\

% \rowcolor{green!10} 
% ~~+ MaskGIL & 75\%  & 200 (\textcolor{orange}{+21.9\%}) &3.96 (\textcolor{red}{-4.21\%}) \\
% \rowcolor{green!10} 
% ~~+ MaskGIL & 50\%  & 136 (\textcolor{orange}{+46.9\%}) &4.23 (\textcolor{red}{-11.31\%})\\
% \rowcolor{green!10} 
% ~~+ MaskGIL & 25\%  & 72 (\textcolor{orange}{+71.9\%}) &4.68 (\textcolor{red}{-23.16\%})\\
% \rowcolor{green!10} 
% ~~+ MaskGIL & 0\%   & 8 (\textcolor{orange}{+96.9\%}) &5.64 (\textcolor{red}{-48.42\%})\\
% \bottomrule
% \end{tabular}
% \label{tab:ar_nar_acc}
% }
% \end{table}

\begin{wraptable}{r}{9cm}
	\setlength{\tabcolsep}{3pt}
        \renewcommand\arraystretch{1.4}
        \caption{Experimental Results of Sampling Steps and FID on ImageNet. ``AR Ratio'' refers to the proportion of sampling steps performed by the AR model relative to the initial total of 256 steps.}
	\resizebox{1.0\linewidth}{!}{%
		\begin{tabular}{l|ccc}
            \toprule
            \bf Method &\bf AR Ratio &\bf Sample Steps &\bf FID$\downarrow$  \\
            \midrule
            
            LlamaGen-L & 100\%  &256 &3.80 \\
            
            \rowcolor{green!10} 
            ~~+ MaskGIL & 75\%  & 200 (\textcolor{orange}{+21.9\%}) &3.96 (\textcolor{red}{-4.21\%}) \\
            \rowcolor{green!10} 
            ~~+ MaskGIL & 50\%  & 136 (\textcolor{orange}{+46.9\%}) &4.23 (\textcolor{red}{-11.31\%})\\
            \rowcolor{green!10} 
            ~~+ MaskGIL & 25\%  & 72 (\textcolor{orange}{+71.9\%}) &4.68 (\textcolor{red}{-23.16\%})\\
            \rowcolor{green!10} 
            ~~+ MaskGIL & 0\%   & 8 (\textcolor{orange}{+96.9\%}) &5.64 (\textcolor{red}{-48.42\%})\\
            \bottomrule
            \end{tabular}
            \label{tab:ar_nar_acc}
        }
\end{wraptable}
\vspace{0.1cm} \noindent \textbf{Evaluation on Class-Driven Image Generation.} We use LlamaGen-L~\citep{llamagen} (343M parameters) as the base class-driven AR model. To clearly demonstrate the effectiveness of our unified framework, we select a smaller MaskGIL-B model (111M parameters) as the MAR model. For evaluation, we measure the FID score on the ImageNet dataset, with the CFG scale set to 2.0. The experimental results are presented in Table~\ref{tab:ar_nar_acc}. We observe that when integrating MaskGIL-B with LlamaGen-L at different AR generation ratios, the FID score increases from 3.80 (for pure LlamaGen-L) to 5.64 (for pure MaskGIL). This trend indicates that while introducing the MAR model accelerates sampling, it comes at the cost of reduced image quality. These findings highlight the need to maintain a fundamental trade-off between sample efficiency and generation quality.

\vspace{0.1cm} \noindent \textbf{Evaluation on Text-Driven Image Generation.} We use Lumina-mGPT~\citep{liu2024lumina} (7B parameters) as the base text-driven AR model, known for its capability to generate flexible and photorealistic images from text descriptions. For the MAR model, we leverage our MaskGIL with 775M parameters. However, since MaskGIL currently supports generation only at a 512$\times$512 resolution, our experiments are constrained to this resolution. The CFG scale for the experiments is set to 4.0. In Figure~\ref{fig:t2i_case2}, we present examples along with inference times for images generated using our unified framework. The results indicate a substantial improvement in inference speed while maintaining high-quality image generation. Specifically, when 25\% of tokens are generated by Lumina-mGPT and the remaining 75\% by MaskGIL, the generation time is reduced by 72.4\%. This effectively mitigates the slow inference speed typically associated with AR models, demonstrating the efficiency of our hybrid approach.

\begin{figure}[!t]
    \centering  \includegraphics[width=0.85\linewidth]{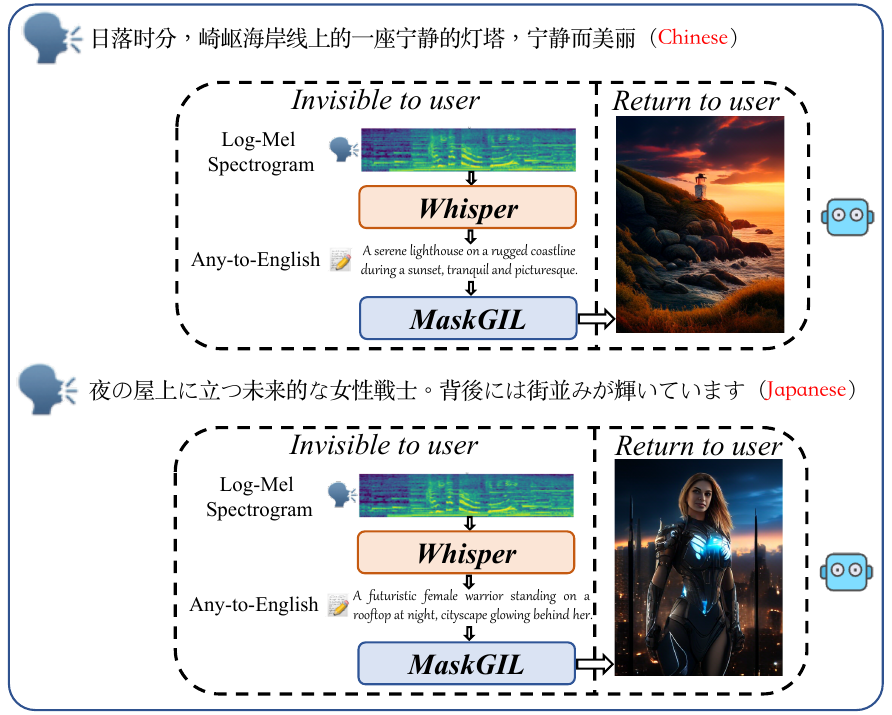}
    \caption{Speech-to-Image Generation System.}
    \label{fig:speech}
\end{figure}

% \begin{figure}[!t]
%     \centering  \includegraphics[width=1.0\linewidth]{figures/eval.jpg}
%     \caption{Speech-To-Image System Evaluation with Image \& Text CLIPScore.}
%     \label{fig:eval_pipeline}
% \end{figure}

\subsection{Real-time Speech-to-Image Generation System}
\noindent \textbf{Any-to-English Speech Translation.} Our text-driven MaskGIL model is trained on an English corpus; however, we aim to enable seamless and intuitive usage for users worldwide. To this end, we develop a speech-to-image generation system that supports over 90 languages, allowing users to generate images simply by speaking. Specifically, the core design leverages Whisper\footnote{Whisper is a general-purpose speech recognition model trained on a large-scale, diverse audio dataset. It is a multitask model capable of both multilingual speech recognition and speech translation.}~\citep{radford2023robust}, a powerful multilingual speech translation model, as an intermediary. The Whisper transcribes the user's spoken input, translates it into English, and then directly utilizes the translated English text as a prompt for image generation, as illustrated in Figure~\ref{fig:speech}.

\vspace{0.1cm}
\noindent \textbf{Enhanced User Interaction.} Since the speech-to-image generation system interacts directly with users, optimizing the user experience is essential. This primarily involves two key aspects: generation speed and image quality. The framework proposed in Section ~\ref{sec:unified} is particularly well-suited for this scenario, as it enables users to balance speed and quality based on their preferences. Specifically, when prioritizing higher image quality, users can increase the contribution of the AR model, accepting potential delays in generation time. Conversely, for faster image synthesis, users can reduce the influence of the AR model, thereby accelerating the generation process.

% \begin{table}[!t]
% \centering
% \setlength{\tabcolsep}{5pt}
% \renewcommand\arraystretch{1.1}
% \caption{Speech-to-Image Generation Performance Evaluated via CLIPScore.}
% \resizebox{0.95\linewidth}{!}{%
% \begin{tabular}{c|cc}
% \toprule
% \bf Method &\bf \makecell{Image \\CLIP Score$\uparrow$} &\bf \makecell{Text \\CLIPScore$\uparrow$}\\
% \midrule
% en text$\to$image & 1.000 &31.040\\
% \rowcolor{green!10} 
% cn audio$\to$image & 0.971 &30.793\\
% \rowcolor{green!10} 
% pt audio$\to$image & 0.986 &30.894\\
% \rowcolor{green!10} 
% fr audio$\to$image & 0.988 &30.891 \\
% \rowcolor{green!10} 
% es audio$\to$image & 0.956 &30.745\\
% \bottomrule
% \end{tabular}
% \label{tab:audio_eval}
% }
% \end{table}

\begin{wraptable}{r}{8.35cm}
	\centering
        \vspace{-0.2cm}
	\setlength{\tabcolsep}{5pt}
        \renewcommand\arraystretch{1.1}
        \caption{Speech-to-Image Generation Performance Evaluated via CLIPScore.}
	\resizebox{0.95\linewidth}{!}{%
            \begin{tabular}{c|cc}
            \toprule
            \bf Method &\bf \makecell{Image \\CLIP Score$\uparrow$} &\bf \makecell{Text \\CLIPScore$\uparrow$}\\
            \midrule
            en text$\to$image & 1.000 &31.040\\
            \rowcolor{green!10} 
            cn audio$\to$image & 0.971 &30.793\\
            \rowcolor{green!10} 
            pt audio$\to$image & 0.986 &30.894\\
            \rowcolor{green!10} 
            fr audio$\to$image & 0.988 &30.891 \\
            \rowcolor{green!10} 
            es audio$\to$image & 0.956 &30.745\\
            \bottomrule
            \end{tabular}
            \label{tab:audio_eval}
        }
\end{wraptable}
\vspace{0.1cm}  \noindent \textbf{Evaluation Data \& Setup.} In order to verify the effectiveness of the system, we construct a text-audio dataset, named PolyVoice\footnote{The dataset are available at~\url{https://github.com/synbol/MaskGIL}}. The dataset is generated in a three-step process: (1) we employ GPT-4o to generate \textbf{English Text$\longleftrightarrow$Multilingual Text} pairs, (2) the multilingual text is synthesized into speech using Google TTS~\citep{gtts}, and (3) we form \textbf{English Text$\longleftrightarrow$Multilingual Audio} pairs. 

The English text used for testing typically ranges from 20 to 50 words in length. Additionally, the audio data encompasses four languages: Chinese (cn), Portuguese (pt), French (fr), and Spanish (es), with each language comprising 100 audio samples. These four languages are selected for testing based on two key factors: 1) their relatively large number of speakers, and 2) their current support by Google TTS.

For evaluation, we leverage CLIPScore~\citep{radford2021learning}, which consists of two key metrics—Image CLIPScore and Text CLIPScore—to measure the accuracy and robustness of the system in speech-driven image generation. Image CLIPScore quantifies the similarity between images generated from audio and those generated from English text, while Text CLIPScore evaluates the semantic alignment between images generated from audio and their corresponding English text descriptions.

\vspace{0.1cm}  \noindent \textbf{Evaluation Results.} The evaluation results in Table~\ref{tab:audio_eval} demonstrate that images generated from multilingual audio inputs maintain strong alignment with those generated from English text. The Image CLIPScore remains high across all languages. Similarly, the Text CLIPScore values exhibit minimal variation, indicating that the system effectively preserves semantic consistency across different languages. These results validate the system’s robustness in cross-lingual speech-to-image generation.

\section{Conclusion}
In this work, we explore the optimal MAR model configuration for efficient and scalable image generation. We develop a diverse set of models for class-driven image synthesis and extend our approach to support text-driven generation. Our class-driven models achieve competitive visual quality compared to widely adopted AR models while demonstrating significant efficiency gains. Simultaneously, our text-driven model preserves high visual fidelity and precise text alignment. Furthermore, we further extend our model to a broader range
of applications, including: (1) accelerating AR-based generation and (2) developing a real-time speech-to-image generation system.

\bibliography{arxiv}

\clearpage
\setcounter{section}{0}
\renewcommand{\thesection}{\Alph{section}}
\counterwithin{table}{section}
\counterwithin{figure}{section}
% \section*{Appendix}
\section{Detailed Evaluation Results of Discrete-Value Image Tokenizer}
\label{appendix:image_tokenizer}
\begin{table*}[!t]
	\centering
	\caption{\textbf{Detailed Experimental Results of Different Image Tokenizers on the AR Paradigm.} The results include FID and IS scores for various CFG settings across 50 to 200 epochs. }
	\label{tab:tokenizer}
	
	\renewcommand\arraystretch{1.25}
	
	\resizebox{0.9\textwidth}{!}{%
		\begin{tabular}{c c | cccccc}
			\toprule
			\multicolumn{1}{l}{} &
			\multicolumn{1}{l}{} &
			\multicolumn{2}{c}{\textit{50 epoch}} &
			\multicolumn{2}{c}{\textit{100 epoch}} &
			\multicolumn{2}{c}{\textit{200 epoch}} \\
			
			\hline
			
			&
			&
			FID &
			IS &
			FID &
			IS &
			FID &
			IS \\
			
			\midrule
			
			&
			CFG = 3.0 &
			9.8846 &
			160.1574 &
			9.1567 &
			170.2633 &
			8.7525 &
			175.5390 \\
			&
			CFG = 2.5 &
			10.2998 &
			135.7815 &
			9.4809 &
			145.6140 &
			9.1254 &
			149.5679 \\
			\multirow{-3}{*}{\textbf{MaskGIT-VQ (1024)}} &
			CFG = 2.0 &
			12.6750 &
			102.6800 &
			11.5306 &
			110.0108 &
			11.2182 &
			114.8622 \\
			
			\hline
			
			&
			CFG = 3.0 &
			8.5799 &
			180.8341 &
			8.1700 &
			193.9580 &
			7.5208 &
			200.0188 \\
			&
			CFG = 2.5 &
			8.9955 &
			153.5894 &
			8.2882 &
			165.1865 &
			7.7708 &
			175.9126 \\

			\multirow{-3}{*}{\textbf{Chameleon-VQ (8192)}} &
			CFG = 2.0 &
			11.6736 &
			115.8210 &
			10.5847 &
			125.7858 &
			9.6992 &
			133.8924 \\
			
			\hline

			&
			CFG = 3.0 &
			9.2867 &
			\textbf{199.3415} &
			8.8726 &
			\textbf{215.3014} &
			8.0206 &
			\textbf{221.3254} \\
			&
			CFG = 2.5 &
			\textbf{8.2499} &
			171.4478 &
			\textbf{7.7843} &
			185.0560 &
			\textbf{6.8314} &
			191.7435 \\
			
			\multirow{-3}{*}{\textbf{LlamaGen-VQ (16384)}} &
			CFG = 2.0 &
			8.7104 &
			133.0047 &
			7.8188 &
			144.1255 &
			7.0980 &
			149.0620 \\
			
			\hline

			&
			CFG = 3.0 &
			10.9688 &
			129.7520 &
			10.8770 &
			  131.0903 &
			9.5125 &
			146.1105 \\
			&
			CFG = 2.5 &
			  15.5837 &
			  96.4929 &
			14.8460 &
			100.0897 &
			14.8755 &
			100.2676 \\

			\multirow{-3}{*}{\textbf{Open-MAGVIT2-VQ (262144)}} &
			CFG = 2.0 &
                25.2163 &
                63.6719 &
			24.1729 &
			66.1162 &
			22.7751 &
			71.5757 \\
			\midrule
		\end{tabular}%
	}
\end{table*}

% \usepackage{multirow}
% \usepackage{graphicx}
\begin{table*}[ht]
	\centering
	\caption{\textbf{Detailed Experimental Results of Different Image Tokenizers on the MAR Paradigm.} The results include FID and IS scores for various CFG settings across 50 to 200 epochs.}
	\label{tab:nar_2}
	
	\renewcommand\arraystretch{2}
	
	\resizebox{\textwidth}{!}{%
		\begin{tabular}{clc | lll | lll | lll | ccc}
			\toprule
			&
			&
			&
			\multicolumn{3}{c}{\textbf{MaskGIT-VQ (1024)}} &
			\multicolumn{3}{c}{\textbf{Chameleon-VQ (8192)}} &
			\multicolumn{3}{c}{\textbf{LlamaGen-VQ (16384)}} &
			\multicolumn{3}{c}{\textbf{Open-MAGVIT2-VQ (262144)}} \\
			
			\midrule
			
			\textbf{} &
			\multicolumn{1}{c}{} &
			&
			\multicolumn{1}{c}{CFG = 3.0} &
			\multicolumn{1}{c}{CFG = 2.5} &
			\multicolumn{1}{c}{CFG = 2.0} &
			\multicolumn{1}{c}{CFG = 3.0} &
			\multicolumn{1}{c}{CFG=2.5} &
			\multicolumn{1}{c}{CFG=2.0} &
			\multicolumn{1}{c}{CFG = 3.0} &
			\multicolumn{1}{c}{CFG = 2.5} &
			\multicolumn{1}{c}{CFG = 2.0} &
			\multicolumn{1}{c}{CFG = 3.0} &
			\multicolumn{1}{c}{CFG = 2.5} &
			\multicolumn{1}{c}{CFG = 2.0}  \\
			
			\midrule
			
			&
			&
			FID &
			16.6349 &
			17.9747 &
			20.2007 &
			16.9894 &
			19.3724 &
			22.9273 &
			15.4235 &
			16.9273 &
			19.2510 &
			90.4830 &
			94.1686 &
			98.6222 \\
			
			&
			\multirow{-2}{*}{\textit{50 epoch}} &
			IS &
			98.2236 &
			89.0695 &
			76.8371 &
			106.1228 &
			92.8579 &
			78.0924 &
			114.4028 &
			100.5291 &
			86.2001 &
			15.1394 &
			14.1799 &
			13.0694 \\
			
			\cline{2-15}
			
			&
			&
			FID &
			12.4982 &
			13.6014 &
			15.5298 &
			11.4843 &
			12.7807 &
			15.2883 &
			10.7954 &
			10.8447 &
			11.6757 &
			57.7810 &
			64.0669 &
			71.2787 \\
			&
			\multirow{-2}{*}{\textit{100 epoch}} &
			IS &
			128.8805 &
			114.9444 &
			99.3552 &
			148.9279 &
			133.6862 &
			112.5623 &
			173.5138 &
			156.9519 &
			136.1681 &
			31.4727 &
			26.7601 &
			22.9549 \\
			
			\cline{2-15}
			
			&
			&
			FID &
			9.9150 &
			10.8601 &
			12.6045 &
			8.5408 &
			9.3071 &
			11.2243 &
			8.9188 &
			8.4767 &
			8.5895 &
			32.8960 &
			43.2282 &
			51.0961 \\
			\multirow{-6}{*}{\textbf{ Bidirectional Transformer}} &
			\multirow{-2}{*}{\textit{200 epoch}} &
			IS &
			154.5364 &
			136.8136 &
			115.7622 &
			189.5333 &
			169.3695 &
			145.7171 &
			213.7065 &
			193.9830 &
			169.5869 &
			64.8568 &
			45.3654 &
			36.6035 \\
			
			\midrule
			
			&
			&
			FID &
			12.5828 &
			13.3734 &
			14.7296 &
			10.6284 &
			  11.3143 &
			12.9145 &
			9.8001 &
			9.7923 &
			10.1693 &
			12.5832 &
			14.9028 &
			18.8675 \\
			&
			\multirow{-2}{*}{\textit{50 epoch}} &
			IS &
			121.7072 &
			109.3891 &
			96.8131 &
			144.3818 &
			130.8738 &
			114.3169 &
			169.3604 &
			152.9462 &
			136.0035 &
			121.8998 &
			104.8236 &
			86.8462 \\
			
			\cline{2-15}
			
			&
			&
			FID &
			9.4469 &
			9.9894 &
			10.9635 &
			8.3184 &
			8.4573 &
			9.4030 &
			8.1221 &
			7.8057 &
			7.7422 &
			9.1883 &
			11.2308 &
			14.6781 \\
			&
			\multirow{-2}{*}{\textit{100 epoch}} &
			IS &
			160.4776 &
			143.2989 &
			126.3950 &
			188.8565 &
			172.2684 &
			152.5737 &
			213.7423 &
			194.7384 &
			174.4664 &
			153.0440 &
			131.4824 &
			108.0728 \\
			
			\cline{2-15}
			
			&
			&
			FID &
			7.9954 &
			8.3240 &
			8.8536 &
			6.7024 &
			6.6765 &
			7.2351 &
			6.9971 &
			6.4922 &
			6.1397 &
			7.1548 &
			8.4162 &
			9.3161 \\
			\multirow{-6}{*}{\textbf{Bidirectional LLaMA}} &
			\multirow{-2}{*}{\textit{200 epoch}} &
			IS &
		    191.4879 &
			173.8534 &
			154.6600 &
			226.5536 &
			206.9128 &
			185.9040 &
			244.6256 &
			222.4675 &
			201.6035 &
			214.9400 &
			172.6521 &
			155.9095 \\
			
			\bottomrule
		\end{tabular}%
	}
\end{table*}

\subsection{Results on Autoregressive Generation}
We present detailed experimental results for various image tokenizers applied to the autoregressive (AR) paradigm in Table~\ref{tab:tokenizer}. These experiments are conducted using the Casual LLaMA architecture and evaluated on the class-conditional ImageNet-1k benchmark. The results include FID and IS scores measured at 50, 100, and 200 epochs under CFG settings of 3.0, 2.5, and 2.0. Among these tokenizers, LLamaGen-VQ with a codebook size of 16,384 outperforms the others. For a comprehensive analysis, please refer to Section~\ref{sec: rethinking} in the main paper.

\subsection{Results on Mask  Autoregressive Generation}
We present detailed experimental results for various image tokenizers applied to the masked autoregressive (MAR) paradigm in Table~\ref{tab:nar_2}. These experiments utilize the Bidirectional LLaMA and Bidirectional Transformer models, evaluated on the class-conditional ImageNet-1k benchmark. The results report FID and IS scores across 50, 100, and 200 epochs under CFG settings of 3.0, 2.5, and 2.0. Among these tokenizers, LlamaGen-VQ demonstrates the best performance for the MAR paradigm. From a model architecture perspective, the Bidirectional LLaMA achieves superior results. For a more detailed analysis, refer to Section~\ref{sec: rethinking} in the main paper.

% \usepackage{multirow}
% \usepackage{graphicx}

\begin{table*}[ht]
	\centering
	\caption{\textbf{Detailed Experimental Results of Scaling MaskGIL.} The results include FID and IS scores for various CFG settings across 50 to 400 epochs.}
	\label{tab:table_scaling}
	
	\renewcommand\arraystretch{2}
	
	\resizebox{\textwidth}{!}{%
		\begin{tabular}{cl | llll | llll | llll | lll}
			\toprule
			\multicolumn{1}{l}{} &
			&
			\multicolumn{4}{c}{\textbf{MaskGIL-B (111M)}} &
			\multicolumn{4}{c}{\textbf{MaskGIL-L (343M)}} &
			\multicolumn{4}{c}{\textbf{MaskGIL-XL (775M)}} &
			\multicolumn{3}{c}{\textbf{MaskGIL-XXL (1.4B)}} \\
			
			\midrule
			
			\multicolumn{1}{l}{} &
			&
			CFG = 3.0 &
			CFG = 2.5 &
			CFG = 2.0 &
			CFG = 1.5 &
			CFG=3.0 &
			CFG=2.5 &
			CFG=2.0 &
			CFG=1.5 &
			CFG=3.0 &
			CFG=2.5 &
			CFG=2.0 &
			CFG=1.5 &
			CFG=3.0 &
			CFG=2.5 &
			CFG=2.0 \\
			
			\midrule
			
			&
			FID &
			9.8001 &
			9.7923 &
			10.1693 &
			11.4326 &
			8.1032 &
			7.3123 &
			6.7532 &
			6.9289 &
			5.9404 &
			5.4523 &
			5.4149 &
			6.4069 &
			4.9994 &
			4.6997 &
			5.0216 \\
			\multirow{-2}{*}{\textit{50 epoch}} &
			IS &
			169.3604 &
			152.9462 &
			136.0035 &
			115.6833 &
			248.7714 &
			229.2365 &
			204.0812 &
			172.9692 &
			259.9695 &
			235.1068 &
			204.2972 &
			172.1452 &
			277.6597 &
			251.9433 &
			218.3809 \\
			
			\cline{2-17}
			
			&
			FID &
			8.1221 &
			7.8057 &
			7.7422 &
			8.2393 &
			6.8974 &
			5.8711 &
			5.3028 &
			5.2657 &
			4.7785 &
			4.2401 &
			4.2148 &
			5.2318 &
			3.8788 &
			4.0727 &
			4.8170 \\
			\multirow{-2}{*}{\textit{100 epoch}} &
			IS &
			213.7423 &
			194.7384 &
			174.4664 &
			150.8941 &
			287.1859 &
			265.8999 &
			235.5590 &
			203.3373 &
			299.0290 &
			271.2699 &
			237.7044 &
			198.0010 &
			285.9541 &
			255.0303 &
			220.0994 \\
			
			\cline{2-17}
			
			&
			FID &
			7.4975 &
			6.4922 &
			6.1397 &
			6.8546 &
			6.2931 &
			5.7252 &
			4.8876 &
			4.5048 &
			4.2346 &
			3.9977 &
			4.1228 &
			  5.5923 &
			3.9649 &
			3.8788 &
			4.4784 \\
			\multirow{-2}{*}{\textit{200 epoch}} &
			IS &
			244.6256 &
			222.4675 &
			201.6035 &
			169.0343 &
			327.0787 &
			303.9240 &
			275.0859 &
			235.0050 &
			314.0859 &
			281.0844 &
			262.0203 &
			205.5776 &
			305.9541 &
			274.0724 &
			240.9620 \\
			
			\cline{2-17}
			
			&
			FID &
			6.9971 &
			6.3837 &
			5.6994 &
			5.6713 &
			6.1342 &
			5.1874 &
			4.5432 &
			4.3718 &
			4.1886 &
			3.9442 &
			4.0512 &
			4.9251 &
			3.8365 &
			3.7199 &
			4.2015 \\
			\multirow{-2}{*}{\textit{300 epoch}} &
			IS &
			271.4360 &
			250.5148 &
			226.8228 &
			194.3450 &
			331.4310 &
			306.8823 &
			272.2563 &
			236.6083 &
			321.1712 &
			291.0879 &
			259.2443 &
			220.4716 &
			306.6679 &
                282.4760 &
			253.4782 \\
			
			\cline{2-17}
			
			&
			FID &
			7.0901 &
			6.2576 &
			5.6450 &
			5.6713 &
			5.4711 &
			4.6372 &
			4.0185 &
			4.0854 &
			4.0931 &
			3.9031 &
			4.0522 &
			4.7214 &
			/ &
			/ &
			/ \\
			\multirow{-2}{*}{\textit{400 epoch}} &
			IS &
			274.9196 &
			255.3847 &
			229.9646 &
			199.2374 &
			336.8414 &
			312.2809 &
			281.1118 &
			239.8479 &
			326.2475 &
			296.2475 &
			269.8136 &
			237.9921 &
			/ &
			/ &
			/ \\
			\bottomrule
		\end{tabular}%
	}
\end{table*}

\begin{table*}[t]
	\centering
	\caption{\textbf{Detailed Experimental Results Under Different Decoding Steps and Different Schedulers.} This experiment shows the checkpoint results of MaskGIL-B at 400 epochs, with CFG$=2.0$.}
	\label{tab:steps}
	
	\renewcommand\arraystretch{1.5}
	
	\resizebox{0.9\textwidth}{!}{%
		\begin{tabular}{c|cc|cc|cc|cc|cc}
			\toprule
			\multicolumn{1}{l|}{} &
			\multicolumn{2}{c|}{4 Steps} &
			\multicolumn{2}{c|}{\textit{8 Steps}} &
			\multicolumn{2}{c|}{\textit{16 Steps}} &
			\multicolumn{2}{c|}{\textit{32 Steps}} &
                \multicolumn{2}{c}{\textit{64 Steps}} \\
			
			\midrule
		      Scheduler 
			&FID
			  &IS
                &FID
			  &IS
                &FID
			  &IS
                &FID
			  &IS
                &FID
			  &IS\\
              \hline
              Root &\textbf{9.4683} &\textbf{177.5355} &7.1539 &\textbf{250.1702} &9.1172 &\textbf{240.3087} &10.3262 &\textbf{225.1267} &10.8516 &\textbf{207.2014}\\
              Linear &10.9076 &162.2468 &6.7640 &239.9988 &8.6355 &233.0458 &9.8178 &216.4662 &10.5585 &204.8842\\
              Square &13.4522 &141.6912 &6.1144 &227.3088 &7.9051 &226.6961 &9.2745 &213.4663 &10.0068 &201.7469\\
              Cosine &12.8029 &145.7202 &6.3108 &230.7038 &8.2024 &228.5306 &9.5711 &214.4882 &10.1930 &201.3121\\
              Arccos &13.2392 &143.9644 &\textbf{5.9056} &220.2331&\textbf{7.3303} &217.5756 &\textbf{8.3202} &205.5842 &\textbf{9.5981} &195.7871\\
            \bottomrule
		\end{tabular}%
	}
\end{table*}

\begin{figure*}[!t]
    \centering
    \includegraphics[width=0.96\linewidth]{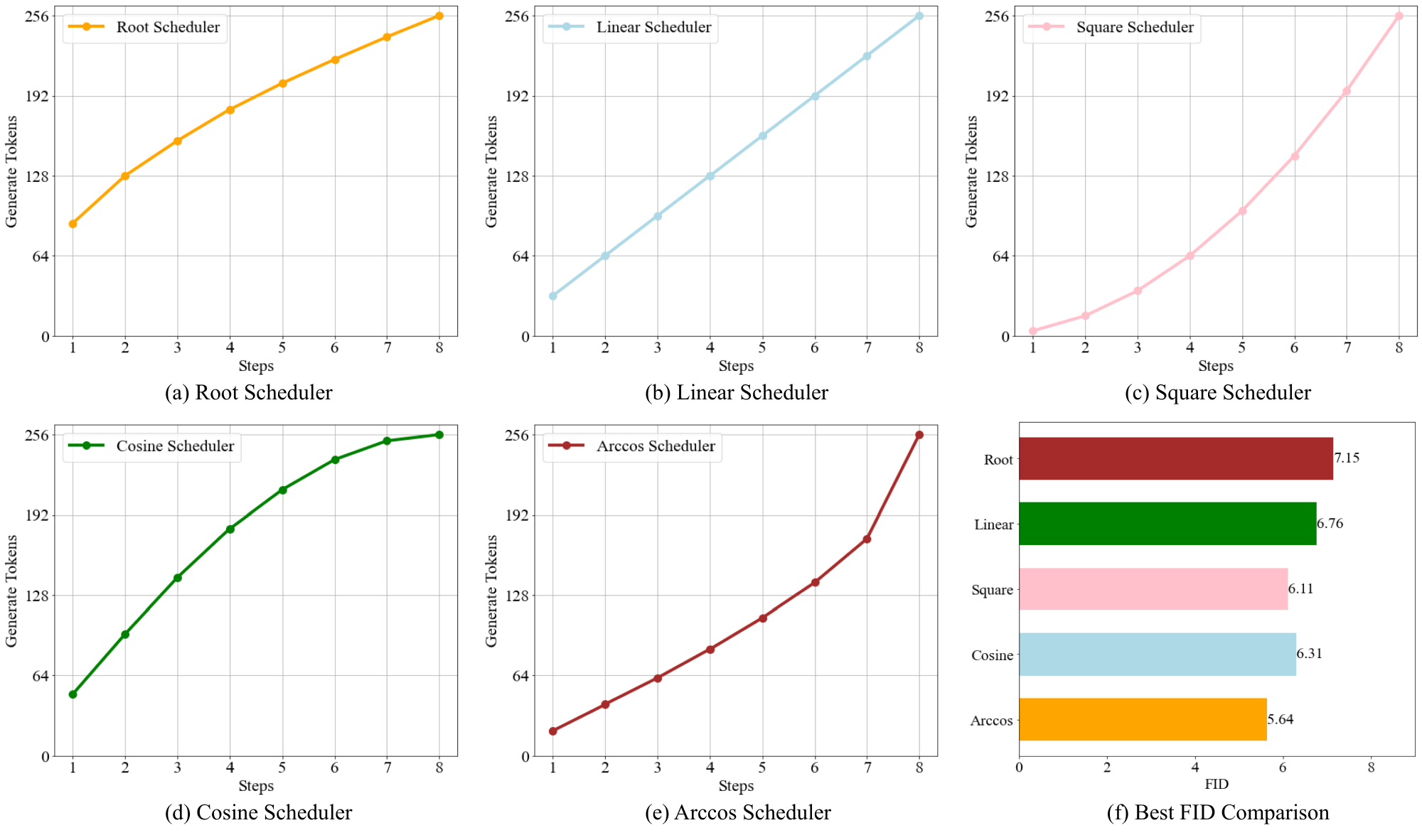}
    \caption{Scheduler Configurations and Metrics on Imagenet 256$\times$256 Benchmark.}
    \label{fig:Scheduler}
\end{figure*}
\section{Details of Mask Schedulers}
\label{sec:sch_appendix}
We evaluate five mask schedulers, including root, linear, square, cosine, and arccos. To provide an intuitive understanding of the differences between these schedulers, we plot the number of generated tokens at each generation step, as illustrated in Figure~\ref{fig:Scheduler}.

\section{Detailed Evaluation Results of Different Decoding Steps}
\label{sec:decoding_steps}
To explore how the number of decoding steps impacts generation quality, we perform an ablation study on the class-conditional ImageNet benchmark, with findings summarized in Table~\ref{tab:steps}. From the results, all mask schedulers present the same conclusion, that is, increasing the number of decoding steps does not consistently enhance generation quality. Notably, 8-step decoding achieves the best results across all mask schedulers.

\section{Detailed Evaluation Results of Scaling MaskGIL}
\label{sec:appendix_scaling}
In this work, we scale the model parameters of MaskGIL from 111M to 1.4B. Table~\ref{tab:table_scaling} details the FID and IS scores for each model size at various training epochs and CFG settings. Notably, the results for our MaskGIL-XXL currently cover only up to 300 epochs. Notable improvements in FID are observed when scaling the model from MaskGIL-B to MaskGIL-XXL. For a detailed analysis, please refer to Section~\ref{sec:4_scaling}.

% \section*{Availability of Data and Materials}
% The ImageNet dataset employed in the class-driven image generation experiment is publicly accessible at~\url{https://www.image-net.org}. The PolyVoice dataset, which was specifically curated to evaluate speech-to-image performance, can be obtained at~\url{https://github.com/synbol/MaskGIL}. However, the training data utilized for our text-driven generation is not publicly available due to potential concerns related to copyright, privacy, and compliance regulations.

% \section*{Author Contributions}
% All authors contributed to the conception and design of the study. Yi Xin and Le Zhuo undertook the tasks of material preparation, data collection, experiment, and analysis. The initial draft of the manuscript was prepared by Yi Xin, Le Zhuo, and Qi Qin, with all authors providing feedback on previous versions of the manuscript. Each author has reviewed and approved the final version of the manuscript.

% \section*{Acknowledgments} This project was supported by the National Natural Science Foundation of China (No. 62301310). The authors would like to thank all
% the anonymous reviewers for their reviews and valuable suggestions. 

% \section*{Declaration of Competing Interest}
% The authors have no competing interests to declare that are relevant to the content of this article.
\end{document}